\documentclass{article}

\usepackage{arxiv}

\usepackage[utf8]{inputenc} % allow utf-8 input
\usepackage[T1]{fontenc}    % use 8-bit T1 fonts
\usepackage{hyperref}       % hyperlinks
\usepackage{url}            % simple URL typesetting
\usepackage{booktabs}       % professional-quality tables
\usepackage{amsfonts}       % blackboard math symbols
\usepackage{nicefrac}       % compact symbols for 1/2, etc.
\usepackage{microtype}      % microtypography
\usepackage{graphicx}
\usepackage{natbib}
\usepackage{doi}
\usepackage[ruled]{algorithm2e}
\usepackage{algorithmic}

\title{Shifting Capsule Networks from the Cloud to the Deep Edge}

\author{ \href{https://orcid.org/0000-0003-2046-4569}{\includegraphics[scale=0.06]{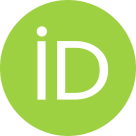}\hspace{1mm}Miguel Costa} \\
    Centro ALGORITMI\\
	Universidade do Minho\\
	Guimarães, 4800-058 \\
	\texttt{miguel.costa@dei.uminho.pt} \\
	\And
	\href{https://orcid.org/0000-0002-0725-5861}{\includegraphics[scale=0.06]{Images/orcid.png}\hspace{1mm}Diogo Costa} \\
	Centro ALGORITMI\\
	Universidade do Minho\\
	Guimarães, 4800-058 \\
	\texttt{diogoandreveigacosta@gmail.com} \\
	\And
	\href{https://orcid.org/0000-0002-4071-9015}{\includegraphics[scale=0.06]{Images/orcid.png}\hspace{1mm}Tiago Gomes} \\
	Centro ALGORITMI\\
	Universidade do Minho\\
	Guimarães, 4800-058 \\
	\texttt{tiago.gomes@dei.uminho.pt} \\
	\And
	\href{https://orcid.org/0000-0003-4580-7484}{\includegraphics[scale=0.06]{Images/orcid.png}\hspace{1mm}Sandro Pinto} \\
	Centro ALGORITMI\\
	Universidade do Minho\\
	Guimarães, 4800-058 \\
	\texttt{sandro.pinto@dei.uminho.pt} \\
}

\renewcommand{\shorttitle}{Shifting Capsule Networks from the Cloud to the Deep Edge}

\hypersetup{
pdftitle={Shifting Capsule Networks from the Cloud to the Deep Edge},
pdfsubject={arxiv article},
pdfauthor={Miguel Costa, Diogo Costa, Tiago Gomes, Sandro Pinto},
pdfkeywords={capsule networks, capsule network quantization, edge, cloud, CMSIS-NN, PULP-NN},
}

\begin{document}
\maketitle

\begin{abstract}
Capsule networks (CapsNets) are an emerging trend in image processing. In contrast to a convolutional neural network, CapsNets are not vulnerable to object deformation, as the relative spatial information of the objects is preserved across the network. However, their complexity is mainly related to the capsule structure and the dynamic routing mechanism, which makes it almost unreasonable to deploy a CapsNet, in its original form, in a resource-constrained device powered by a small microcontroller (MCU). In an era where intelligence is rapidly shifting from the cloud to the edge, this high complexity imposes serious challenges to the adoption of CapsNets at the very edge. To tackle this issue, we present an API for the execution of quantized CapsNets in Arm Cortex-M and RISC-V MCUs. Our software kernels extend the Arm CMSIS-NN and RISC-V PULP-NN to support capsule operations with 8-bit integers as operands. Along with it, we propose a framework to perform post-training quantization of a CapsNet. Results show a reduction in memory footprint of almost 75\%, with accuracy loss ranging from 0.07\% to 0.18\%. In terms of throughput, our Arm Cortex-M API enables the execution of primary capsule and capsule layers with medium-sized kernels in just 119.94 and 90.60 milliseconds (ms), respectively (STM32H755ZIT6U, Cortex-M7 @ 480 MHz). For the GAP-8 SoC (RISC-V RV32IMCXpulp @ 170 MHz), the latency drops to 7.02 and 38.03 ms, respectively.
\end{abstract}

% keywords can be removed

\section{Introduction}
%% Why Capsule Networks?
Convolutional layers have shown an impressive capacity to extract features - edges and shapes - from image pixels \citep{Sabour2017, Mensah2019, Omaima2021}. However, to reduce the dimension of feature maps, and consequently the number of parameters to learn, convolutional layers are typically connected through pooling layers in a convolutional neural network (CNN). Pooling layers take the most important features present in a region of the feature map generated by a convolution and pass those features to the following layer. However, this comes at the cost of decreased capacity to recognize pose and object deformation as the relative spatial information of the features is lost across the network \citep{Sabour2017, Mensah2019, Omaima2021}. Furthermore, CNNs have shown to be vulnerable to adversarial examples, i.e., inputs to a machine learning (ML) model that an attacker intentionally designs to compromise the integrity of the decision in a process similar to an optical illusion \citep{Su2019, Sadeghi2020}. To tackle these issues, \citet{Sabour2017} proposed a new class of deep learning (DL) architectures, named capsule neural networks (CapsNets), to extract and classify features from images, while preserving and learning spatial information.

%% What is a Capsule Network and how does it surpass CNNs?
The core unit of a CapsNet is a capsule, a group of neurons organized as vectors. Each vector dimension represents a different parameter of the same object in the image \citep{Sabour2017} and the length of the vector represents the probability of the image containing the object. To date, there are four main variants of CapsNet: (i) transforming auto-encoders \citep{Hinton2011}, (ii) routing by agreement between capsules \citep{Sabour2017}, (iii) expectation-maximization routing capsules \citep{Geoffrey2018}, and (iv) stacked capsule autoencoders \citep{stacked_caps}. However, the most prominent remains the routing by agreement between capsules \citep{Sabour2017}, which is the target on this work. Dynamic routing is used to calculate a series of coefficients that determine the probability of an object in a lower capsule layer, being part of an object represented in a higher capsule layer \citep{Sabour2017}. In contrast to CNNs, CapsNets are equivariant, learning proportion and pose changing \citep{Omaima2021}. CapsNets already outperform CNNs in some healthcare \citep{Chaos2020} and automation mechanisms \citep{Amara2018, Pari2019, Wang2018, Katebi2019, Mensah2019, Zhang2020}.

%% Drawbacks of Capsule Networks in terms of computational requirements
Despite being promoted as the next ground-breaking algorithm in DL \citep{Mensah2019}, CapsNets in their original form are computationally intensive in terms of memory requirements (bandwidth and size) and latency \citep{Marchisio2020, Marchisio2020_2, Marchisio2021,  Zhang2020, Zhang2021, Park2020, Kakillioglu2020}. For the same dataset, the original CapsNet has a ratio of multiply-accumulate (MAC) operations per memory near 100x higher than AlexNet \citep{Marchisio2020}, which is also a heavy CNN. In an era where concerns with cloud and internet infrastructure are disrupting the shift of AI to the deep edge \citep{Stoica2017, CMSIS-NN, Vestias2020, Li2018}, the feasibility of CapsNets is still mainly confined to powerful cloud servers. To shift AI to the edge, researchers have been developing mechanisms for compressing artificial neural networks (ANNs) and CNNs, discarding floating-point representation, and adopting a lighter fixed-point representation \citep{CMSIS-NN, PULP-NN}. If the potential of fixed-point computation in ANNs and CNNs is already known and ultimately led to the development of APIs tailored for the execution of quantized neural networks in low-end MCUs \citep{CMSIS-NN, PULP-NN} (e.g., Arm Cortex-M and RISC-V), the feasibility of CapsNets under such constraints is still unknown.

%% Contributions
To fill this gap, we evaluate the feasibility of CapsNets on resource-constrained platforms, powered by small MCUs targeted for intelligent IoT edge nodes. For this purpose, we extended the software kernels from CMSIS-NN \citep{CMSIS-NN} and PULP-NN \citep{PULP-NN} to support the deployment of CapsNets on Arm Cortex-M and RISC-V MCUs, respectively. We extended these APIs with functions to support capsule operations with int-8 data and evaluated their feasibility in terms of latency, accuracy penalty, and memory footprint reduction. As ML models are typically trained using floating-point precision, we developed a toolchain to perform post-training quantization of CapsNets developed in TensorFlow. Results show that this framework can reduce the memory footprint of a CapsNet by almost 75\%, with a maximum accuracy loss of 0.18\%. Our API was tested in three different Arm Cortex-M cores, i.e., STM32L4R5ZIT6U (Cortex-M4 @ 120 MHz), STM32H755ZIT6U (Cortex-M7 @ 480 Mhz), and STM32L552ZET6QU (Cortex-M33 @ 110 MHz). For the fastest MCU (STM32L4R5ZIT6U), our API enables the execution of primary capsule and capsule layers with medium-sized kernels in just 119.94 and 90.60 milliseconds, respectively. The API for the RISC-V RV32IMCXpulp architecture was tested on a Gapuino v1 board. For an octa-core setting at 170 MHz, the latency of medium-sized primary capsule and capsule layers dropped to 7.02 and 38.03 milliseconds, respectively. Our work is open-source and available online \footnote{\url{https://gitlab.com/ESRGv3/q7-capsnets.git}}. We believe our work will open a path in the application space of CapsNets, mainly in areas till now only explored by CNNs \citep{Costa2019, Lu2021} and encourage the development of lighter CapsNet architectures.

In summary, we make the following contributions: (i) the first public and open-source software kernels for the execution of CapsNets in Arm Cortex-M and RISC-V (RV32IMCXpulp) MCUs (section \ref{sec:sw_kernels}); (ii) a public and open-source framework to quantize CapsNets from floating-point precision to int-8, compliant with the $Qm.n$ data format, used by Arm CMSIS-NN and RISC-V PULP-NN libraries (section \ref{sec:quant_fw}); (iii) the first public evaluation of the feasibility of CapsNets in resource-constrained MCUs featuring Arm Cortex-M and RISC-V (RV32IMCXpulp) cores (section \ref{sec:results}).

\newpage
\section{Background}
\subsection{Capsule Networks (CapsNets): Overview and Evolution} \label{sec:capsnets_overview}

\begin{figure}[b]
	\centering
	\includegraphics[width=0.8\linewidth]{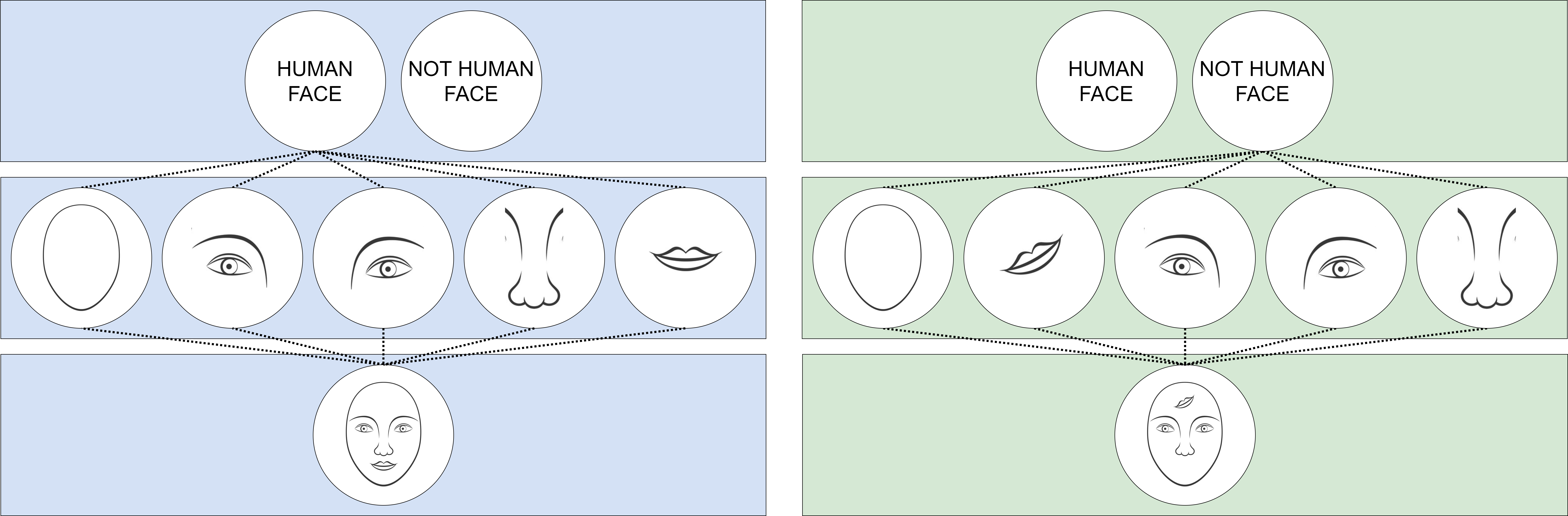}
	\caption{Basic intuition behind CapsNets}
	\label{fig:CapsNet_Basic}
\end{figure}

The first concept of capsules applied in DL was introduced in 2011 by \citet{Hinton2011} to tackle the lack of equivariance affecting CNNs and to show how neural networks can be used to recognize pose. Capsules were described as sets of neurons, organized as vectors, and each capsule represented a given object of the input image. In contrast to a CNN, in which layers take as input/output scalars, capsules deal with vectors. They encapsulate posture information (position, direction, etc.) with other instantiation parameters (width, height, etc.) that together represent an object or object part. Each layer in a CapsNet is made up of a series of capsules and CapsNets follow a part-whole hierarchy. Capsules at one layer predict larger objects, represented by capsules at higher layers, via transformation matrices applied to their output vectors. For an object to be detected, capsules at a lower layer must agree in their spatial arrangement. This characteristic is highlighted in Figure \ref{fig:CapsNet_Basic}. As observed, for the CapsNet in green, the pose of the mouth is not spatially aligned with the remaining parts of the face, resulting in a distorted human face to the human eye. As CapsNets learn the spatial relation of object parts, they can easily spot this distortion. In contrast, a traditional CNN would recognize the input image as a human face as CNNs learn the parts but not their spatial arrangement. Researchers have been proposing numerous CapsNet architectures since the proposal of the capsule concept in 2011. Here we review the most relevant architectures developed to date. For an overview of other architectures, we refer to works \citep{li2021, Omaima2021, Mensah2019, zengguo2022}.

Transforming auto-encoders (TAEs) are the first CapsNet architecture \citep{Hinton2011}. They were designed to encode the input image and regenerate it in the same original pose. TAEs differed from other auto-encoders as they were designed to explicitly learn the transformation matrix that represents the relation of one part of an image (e.g. eyes) to the position of other parts (e.g. nose). A TAE's capsule is composed of two distinct units: (i) recognition units and (ii) generative units. The recognition units were designed to compute the pose parameters and the probability that the object represented by the capsule is present in the input image. The generative units take this data as input to reconstruct the object. The major bottleneck of TAEs is that they require the transformation matrix to be supplied externally.

The previous capsule structure was updated in 2017 for classification tasks. \citet{Sabour2017} succeeded in implementing a CapsNet with state-of-the-art accuracy of 99.75\% on MNIST and triggered the hype around CapsNets. This architecture features a convolutional layer, a primary capsule layer, and a class capsule layer. The convolutional layer is intended to extract the most relevant features of the input image, tackling the need to externally provide the pose matrix of the objects of interest. The primary capsule layer is the first capsule layer following the convolutional layer. Primary capsules have a convolutional kernel and encode the instantiation parameters of the lower-level objects of the input image. The last layer is the class capsule layer, which makes the final prediction. Capsules at different layers connect through a dynamic routing algorithm, which computes the probability of the object represented by a capsule in layer $L$ being part of the objects represented by the capsules in layer $L+1$. \citet{Sabour2017} updated the capsule structure in this CapsNet architecture. Capsules are still a group of neurons that output a vector with instantiation parameters. However, \citet{Sabour2017} gave meaning to the length of the vector - it represents the probability of the object being present in the input image. This CapsNet architecture is the most supported by the research community and is the focus of this work. Section \ref{sec:capsnets_dr} describes it in detail.

In 2018, \citet{Geoffrey2018} proposed to replace the input and output vectors of capsules with matrices and updated the dynamic routing with an expectation-maximization algorithm. Instead of a pose vector, each capsule comprises a $4x4$ pose matrix and an activation parameter that represents the probability of the object is present in the input image. This tackles two of the biggest bottlenecks of the previous CapsNet architecture. Firstly, representing capsules with matrices of $n$ elements instead of vectors reduces the number of parameters in transformation matrices between capsules from $n^2$ to $n$. Secondly, it avoids the use of a \textit{squash} function to shrink the vectors to a maximum length of 1, which prevents the existence of any sensible objective function in the network. The new dynamic routing considers every capsule in a higher layer represented by a Gaussian distribution, defined by a mean and a standard deviation, and the capsules in a lower layer as the data points to be mapped to the Gaussian distributions. This algorithm encompasses two phases: (i) the assignment of data points to clusters (expectation) and (ii) the update of the centroids giving the assignments (maximization). This architecture achieved an accuracy of 98.2\% on the smallNORB dataset.

The previous CapsNet architecture was updated in 2019 by \citet{stacked_caps} for unsupervised learning applications. In addition to the pose matrix and activation parameter, this newer version of capsules (stacked capsule autoencoders) features a vector that represents other properties than pose, such as deformation, velocity, or color. While previous versions of capsules predict the pose of a parent capsule from the pose of the lower-layer capsules, this new architecture works in the opposite direction. A capsule that represents an object at one layer, predicts the pose of the parts represented in the lower layers. Consequently, there is no more need for iterative routing at inference time as every part is explained as a mixture of predictions from different objects. Stacked capsule autoencoders feature two main parts: (i) part capsule autoencoder and (ii) object capsule autoencoder. Stacked capsule autoencoders achieved a state-of-the-art accuracy for unsupervised classification on SVHN (55\%) and MNIST (98.7\%).

%%%%%%%%%%%%%%%%%%%%%%%%%%%%%%%%%%%%%%%%%%%%%%%%%%%%%%%%%%%%%%%%%%%%%%%%%%%%%%%%%%%%%%%%%%%%%%
%%%%%%%%%%%%%%%%%%%%%%%%%%%%%%%%%%%%%%%%%%%%%%%%%%%%%%%%%%%%%%%%%%%%%%%%%%%%%%%%%%%%%%%%%%%%%%
%%%%%%%%%%%%%%%%%%%%%%%%%%%%%%%%%%%%%%%%%%%%%%%%%%%%%%%%%%%%%%%%%%%%%%%%%%%%%%%%%%%%%%%%%%%%%%
\subsection{Capsule Networks (CapsNets) with Dynamic Routing} \label{sec:capsnets_dr}

\begin{figure}[h]
	\centering
	\includegraphics[width=0.75\linewidth]{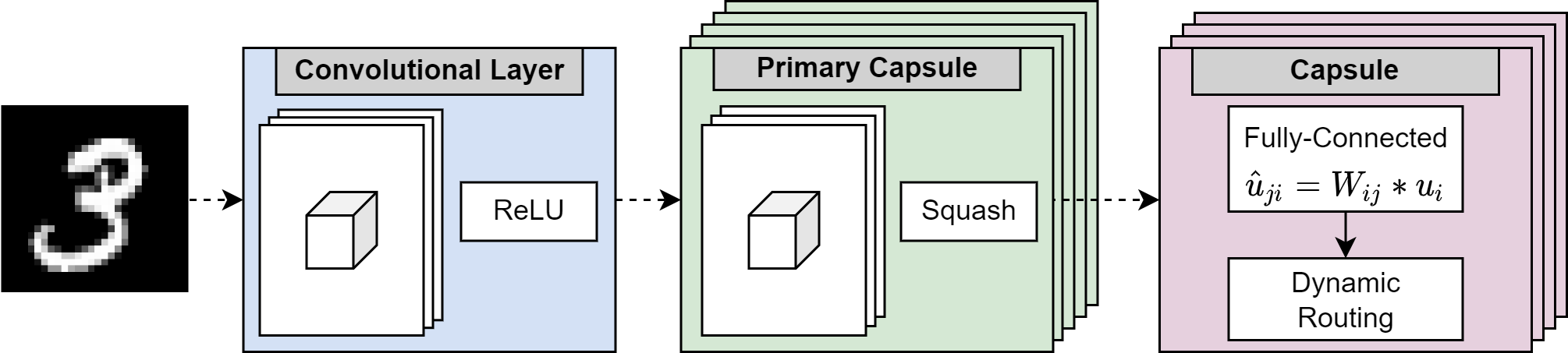}
	\caption{General architecture of a CapsNet with dynamic routing between capsules}
	\label{fig:CapsNet_Arch}
\end{figure}

Dynamic routing between capsules was the first CapsNet architecture proposed for classification tasks. Its successor, matrix capsules with expectation-maximization routing, promised to reduce the number of parameters to learn and to deliver better results in more complex datasets like smallNORB. Nevertheless, at the time of accomplishment of this work, the first version still got the most valuable support from the research community. Furthermore, it is the only version whose original implementation is publicly available \footnote{https://github.com/Sarasra/models/tree/master/research/capsules}. Figure \ref{fig:CapsNet_Arch} depicts its general architecture.

A CapsNet based on dynamic routing between capsules starts with a convolutional layer, used to extract the most relevant features of the input image. In their model tuned for the MNIST dataset, \citet{Sabour2017} employed 256 convolution kernels with dimensions 9x9 and a stride of 1, followed by a \textit{rectified linear unit (ReLU)} activation. The activated feature map is then sent to every capsule in the primary capsule layer. The primary capsule layer is the first capsule layer of the network and is where the inverse of the rendering process starts. Primary capsules have a convolutional kernel and are designed to refine the objects of interest identified by the previous convolutional layer and to determine the instantiation parameters, including spatial information. Each primary capsule represents a different object from the lowest level objects of the input image. The output of a primary capsule is a vector, where each dimension represents a given parameter of the object. The length of the output vector represents the probability of the object is present in the input image. For this purpose, the output of each capsule is squashed by the non-linear activation detailed in Equation \ref{eq:squash_func}. It preserves the original orientation of the output vector but normalizes its length to the range [0-1]. In their model tuned for the MNIST dataset, \citet{Sabour2017} propose a primary capsule layer with 32 convolutional capsules, each one with 8-dimensional features. The convolution relies on a 9x9 kernel and a stride of 2.

\begin{equation}
    v_j = \frac{||s_j||^{2}}{1+||s_j||^{^{2}}} \frac{s_j}{||s_j||}
    \label{eq:squash_func}
\end{equation}

The primary capsule layer is followed by a single or multiple capsule layer. In the last capsule layer, each capsule represents a different object/class to be recognized by the network. For all but the primary capsules, the input is a weighted sum of the output vectors from the capsules in the layer below. For a given capsule $j$ in layer $L+1$, each output vector $u_i$ from layer $L$ is multiplied by the respective weight matrix $W_{ij}$ as detailed in Equation \ref{eq:caps_input}. The resulting vector $\hat{u}_{ij}$ (prediction vector) is submitted to the iterative dynamic routing mechanism. This mechanism computes the coupling coefficients $c_{ij}$ between each capsule in layer $L$ and capsules in layer $L+1$. Coupling coefficients determine the probability of the object represented by a capsule in layer $L$ being part of a capsule in layer $L+1$.

\begin{equation}
    \label{eq:caps_input}
    s_j = \sum_{i}^{}c_{ij} * \hat{u}_{ji},\;\;\; \hat{u}_{ji} = W_{ij} * u_i 
\end{equation}

Algorithm \ref{alg:dynamic_rout} details the dynamic routing algorithm \citep{Sabour2017}. In the first iteration of the routing process, the prediction vector of a capsule $i$ in layer $L$ is equally routed to every capsule $j$ of layer $L+1$. This occurs as logits $b_{ij}$ of the coupling coefficients are initialized as zeros. In the next iterations, each coupling coefficient $c_{ij}$ is refined by measuring the agreement between the prediction vector of capsule $i$ and the output of capsule $j$. The agreement is calculated as the scalar product between the prediction vector $\hat{u}_{ij}$ and the output vector $v_j$. This agreement is added to the initial logits $b_{ij}$. The output of a capsule at a given iteration is computed as the scalar product of the prediction vectors and the respective coupling coefficients. To ensure that the coupling coefficients between a capsule of layer $L$ and all capsules from layer $L+1$ sum to 1, the logits $b_{ij}$ that reflect the level of agreement between capsules are submitted to a \textit{softmax} function before the output vector is calculated. As the output vector is calculated, it is submitted to a \textit{squash} function to ensure that its length belongs to the interval [0, 1]. The routing algorithm is repeated for a given number of iterations, previously defined or hyper-tuned by the neural network designer. In their network specifically tuned for the MNIST dataset, \citet{Sabour2017} uses a unique capsule layer, composed of 10 16-dimensional capsules.

\begin{algorithm}[h]
\small
\caption{Dynamic routing algorithm}
\label{alg:dynamic_rout}
\begin{algorithmic}[1]
\STATE for all capsule $i$ in layer $L$ and capsule $j$ in layer $L+1$: $b_{ij} = 0$
\FOR{$r$ iterations}
    \STATE for all capsule $i$ in layer $L$: $c_i = softmax(b_i)$ 
    \STATE for all capsule $j$ in layer $L+1$: $s_j = \sum_{i}^{}c_{ij} * \hat{u}_{ji}$
    \STATE for all capsule $j$ in layer $L+1$: $v_j = squash(s_j)$
    \STATE for all capsule $i$ in layer $L$ and capsule $j$ in layer $L+1$: $b_{ij} = b_{ij} + \hat{u}_{ji} * v_j$
\ENDFOR
\RETURN $v_j$
\end{algorithmic}
\end{algorithm}

%%%%%%%%%%%%%%%%%%%%%%%%%%%%%%%%%%%%%%%%%%%%%%%%%%%%%%%%%%%%%%%%%%%%%%%%%%%%%%%%%%%%%%%%%%%%%%
%%%%%%%%%%%%%%%%%%%%%%%%%%%%%%%%%%%%%%%%%%%%%%%%%%%%%%%%%%%%%%%%%%%%%%%%%%%%%%%%%%%%%%%%%%%%%%
%%%%%%%%%%%%%%%%%%%%%%%%%%%%%%%%%%%%%%%%%%%%%%%%%%%%%%%%%%%%%%%%%%%%%%%%%%%%%%%%%%%%%%%%%%%%%%

\subsection{Quantization}
Quantization refers to techniques for mapping floating-point tensors to integer tensors, with lower bit-width \citep{Amir2021, Hao2020, Liang2021, Novac2021, Wang2020}. As a consequence, a quantization process involves the encoding of (i) the sign, (ii) the integer part, and (iii) the fractional part of a float in a single integer value. In binary, quantized numbers are typically represented in $Qm.n$ notation, where $m$ specifies the number of bits allocated for the integer part and $n$ the number of bits allocated for the fractional part \citep{Novac2021}. The main goal of quantization is to improve inference latency and reduce the memory footprint of neural networks. However, the success of a quantization procedure is highly dependent on the memory specification and instruction set architecture (ISA) of the target platform \citep{Amir2021, Novac2021}.

\paragraph{\textbf{Post-training quantization vs. quantization-aware training:}} Quantization-aware training lowers the impact of the quantization on model accuracy by simulating the low-precision of weights during training. It considers the quantization error as part of the loss returned by the loss function, which the optimization algorithm tries to minimize \citep{Amir2021, Hao2020, Liang2021, Novac2021, Wang2020}. In post-training quantization, the model is quantized without any fine-tuning during training.

\paragraph{\textbf{Uniform vs. non-uniform quantization:}} Quantizing floating-point values to lower bit-width integers can return values that are equally (uniform quantization) or non-equally (non-uniform quantization) spaced \citep{Amir2021, Novac2021, Wang2020}. For a fixed bit-width, non-uniform quantization may return more accurate ML models, as it captures with higher resolution the more important value regions \citep{Amir2021, Novac2021}. The quantization step is tuned to minimize the quantization error \citep{Amir2021, Novac2021, Wang2020}. Nevertheless, non-uniform quantization typically introduces non-negligible overhead in the computation of quantized models on general-purpose hardware (GPUs and CPUs), sometimes leading to a latency higher than using the floating-point ML model \citep{Novac2021, Wang2020}. As a consequence, uniform quantization is the de-facto method to perform fast computations on resource-constrained devices \citep{Amir2021, Novac2021}. Equation \ref{eq:quant_func} represents a typical uniform quantization function, where $r$ is the float to be quantized, $S$ is a scalar, and $Z$ is a zero-point or offset.

\begin{equation}
    \label{eq:quant_func}
    Q(r) = round(r/S) - Z
\end{equation}

\paragraph{\textbf{Scaling and offset:}} In uniform quantization, the scaling factor and the zero-point are the core of the quantization function \citep{Amir2021, Novac2021, Hao2020, Jacob2018}. The scaling factor specifies the step size of the quantizer, dividing a given range of values into equally spaced partitions. The zero-point is an integer offset used to ensure that a zero is quantized with no error when the range of values to be quantized is not centered around 0. To define a scaler (Equation \ref{eq:scaler}), it is required to firstly define the target $Qm.n$ format and the range of floating-point values to be quantized ($[x_{min}, x_{max}]$) \citep{Amir2021, Hao2020}. The clipping range can be symmetric ($|x_{min}| = |x_{max}|$) or asymmetric ($|x_{min}| != |x_{max}|)$). For an asymmetric setting, the zero-point is calculated as defined in Equation \ref{eq:zero_point} \citep{Hao2020}. Note that the zero-point is rounded to an integer value. Asymmetric quantization often results in a tight clipping range, which can reduce the impact of quantization on model accuracy \citep{Amir2021, Novac2021}. Nevertheless, symmetric quantization is widely adopted in practice as asymmetric quantization can introduce non-negligible overhead during inference \citep{Amir2021}.

\begin{equation}
    \label{eq:scaler}
    S = \frac{x_{max} - x_{min}}{2^{b} - 1}
\end{equation}

\begin{equation}
    \label{eq:zero_point}
    z = -round(x_{min} * S) - 2^{b}-1
\end{equation}

\paragraph{\textbf{Quantization granularity:}} Two metrics should be considered when selecting the granularity of quantization: (i) impact on model accuracy and (ii) computational cost \citep{Hao2020}. The minimum impact on model accuracy is achieved if the scaling factor and the zero-point are calculated for each weight and activation. However, this would result in a tremendous computation overhead and memory consumption as scaling and zero-point factors must be saved for each weight/activation. These factors combined may hamper the main objective of the quantization \citep{Novac2021, Amir2021}. In contrast, the minimum computational cost is achieved if the scaling factor and the zero-point are calculated for the whole neural network \citep{Novac2021, Amir2021}. However, this may result in an accuracy drop that can not be neglected. As a consequence, performing the quantization layer-by-layer or filter-by-filter are the most common strategies \citep{Novac2021, Wang2020}.

\paragraph{\textbf{Fixed bit-width vs. mixed bit-width:}} In a fixed bit-width setting, the same bit-width is used to quantize the whole network \citep{Amir2021, Wang2020}. In a mixed bit-width setting, the bit-width of each layer or filter is tuned for better model accuracy and/or memory footprint reduction \citep{Amir2021, Wang2020}. Activations and weights for layers/filters whose contribution to the final output is smaller get lower bit-width, while layers/filters that contribute the most get higher bit-widths. For the same memory footprint, this strategy has the potential to reduce the degradation of the model accuracy, however at the cost of a more complex quantization scheme and inference-pass software \citep{Amir2021, Wang2020}.

\paragraph{\textbf{Static vs. dynamic quantization:}} In static quantization, the scaler and the zero-point factors are calculated before inference, using a representative dataset. In dynamic quantization, these parameters are computed dynamically during the inference pass for each input \citep{Amir2021, Novac2021}. In this approach, weights and activations are stored in memory in low-precision bit-width, but the operations (matrix multiplication and convolution) are performed in floating-point arithmetic. In comparison to static quantization, dynamic quantization lowers the impact of quantization on model accuracy; however, it incurs a performance overhead that may be prohibited for low-end MCUs \citep{Amir2021, Novac2021}.

%%%%%%%%%%%%%%%%%%%%%%%%%%%%%%%%%%%%%%%%%%%%%%%%%%%%%%%%%%%%%%%%%%%%%%%%%%%%%%%%%%%%%%%%%%%%%%
%%%%%%%%%%%%%%%%%%%%%%%%%%%%%%%%%%%%%%%%%%%%%%%%%%%%%%%%%%%%%%%%%%%%%%%%%%%%%%%%%%%%%%%%%%%%%%
%%%%%%%%%%%%%%%%%%%%%%%%%%%%%%%%%%%%%%%%%%%%%%%%%%%%%%%%%%%%%%%%%%%%%%%%%%%%%%%%%%%%%%%%%%%%%%

\subsection{CMSIS-NN}
CMSIS-NN \citep{CMSIS-NN} is a series of open-source software kernels optimized to enable the deployment of neural networks on resource-constrained Arm Cortex-M MCUs. These software kernels are built upon the CMSIS abstraction layer and are optimized for Cortex-M MCUs that support single instruction multiple data (SIMD). This API is organized in two parts: \textit{NNFunctions} and \textit{NNSupportFunctions}. \textit{NNFunctions} include functions for the computation of the most popular layer types, such as convolution and fully-connected. \textit{NNSupportFunctions} include support functions for data precision conversion and activations, which are used in the computation of \textit{NNFunctions}. These kernels support both 8-bit and 16-bit data. The quantization is assumed to be uniform, symmetric, and to return values with fixed bit-width. The quantization must also be static and conducted layer-by-layer. As the quantization relies on a power-of-two scaling, the scaling of a layer output is implemented as a bitwise shift operation.

For a convolutional layer, input data is assumed to be in Height-Width-Channel (HWC) format, i.e., channel first. In terms of activation functions, CMSIS-NN supports \textit{ReLU}, \textit{sigmoid}, and \textit{tanh} \citep{Dong2020}. \textit{ReLU} is implemented as a simple clipping function that maps negative values to 0 and maintains the value of positive values. \textit{Sigmoid} and \textit{tanh} are implemented using a lookup table (LUT) approach. These kernels were firstly tested by \citet{CMSIS-NN} on the STM32F746ZG MCU (Arm Cortex-M7) running a CNN featuring three convolutional layers and one fully-connected layer, at 216 MHz. CMSIS-NN was able to classify 10.1 images per second with an accuracy drop of only 0.04\%.

\subsection{PULP-NN}
PULP-NN \citep{PULP-NN} is an open-source API optimized for the execution of quantized neural networks in RISC-V RV32IMCXpulp MCUs. As this API is based on CMSIS-NN dataflow, it is also divided into two parts: \textit{NNKernels} and \textit{NNUtils}. \textit{NNKernels} implement matrix multiplication and the standard layer types of neural networks, while \textit{NNUtils} include support functions for data precision conversion and batch normalization. PULP-NN supports 8-bit and sub-byte (4-bit, 2-bit, and 1-bit) data types. However, in terms of activation functions its offering is more limited as it only supports \textit{ReLU}. For the 8-bit setting, the quantization of weights is processed as for CMSIS-NN. Notwithstanding, the quantization of activations follows a different approach, as PULP-NN only supports non-negative activations. Consequently, the quantization must be asymmetric. Similar to CMSIS-NN, data is assumed to be in the HWC layout.

The key innovation of PULP-NN is the support for multi-core processing. For a convolution, the workload can be split along the height, width, or channel dimension of the output feature map. For a fully-connected layer, the workload is split along the output neurons. This library was firstly tested by \citet{PULP-NN} in a GAP-8 MCU executing a CNN quantized to int-8 data and featuring three convolutional layers and one fully-connected layer. In an octa-core setup at 170 MHz, PULP-NN enables this CNN to run in 30x and 19.6x fewer clock cycles than the CMSIS-NN library, running on STM32L476 (Armv7-M, Cortex-M4)  and STM32H743 (Armv7-M, Cortex-M7) MCUs, respectively.

\section{Software Kernels} \label{sec:sw_kernels}
\subsection{Matrix Multiplication} \label{sec:mat_mul}
Matrix multiplication is a memory-intensive operation. Consequently, reducing memory access by maximizing data reuse at the register file level is a key factor in increasing the throughput of CapsNets. As Armv7E-M, Armv8-M, and RISC-V RV32IMCXpulp are 32-bit computing architectures and our software kernels use 8-bit data as operands, matrix multiplication can strongly benefit from a SIMD to read four operands from memory at once. The maximum throughput would be achieved by combining this SIMD with a SIMD for the simultaneous MAC of two words containing four 8-bit operands (4x8-bit word). Since the ISAs of Armv7E-M and Armv8-M are very distinct from the ISA of RISC-V (RV32IMCXpulp), the matrix multiplication kernel was tuned for each computing architecture.

\subsubsection{\textbf{Arm Cortex-M}} \label{sec:MM_ARM}
CMSIS-NN already provides a function for the multiplication of int-8 matrices: \textit{arm\_mat\_mult\_q7}. However, for all but the latest ARMv8.1-M architecture, this function does not rely on any type of hardware acceleration, including SIMD instructions, to read from memory, and for MAC operations. It simply iterates over the rows of the source matrix $A$ and over the columns of the source matrix $B$, reading one element at a time and multiplying them. It uses a 32-bit accumulator, which saturates at 8-bit after the proper right shifting determined in the quantization process. \textit{arm\_mat\_mult\_q7} does not use any loop unrolling technique or parallelism. For a 4x4 sized matrix multiplication kernel, this function requires 8 load operations without sign extension and 4 MACs. As matrix multiplication is a very critical operation within the capsule computation, we tried to optimize the CMSIS-NN baseline function and developed two additional functions for quantized matrix multiplication: (i) \textit{mat\_mult\_q7\_trb} and (ii) \textit{mat\_mult\_q7\_simd}.

\begin{figure}[t]
	\centering
	\includegraphics[width=\linewidth]{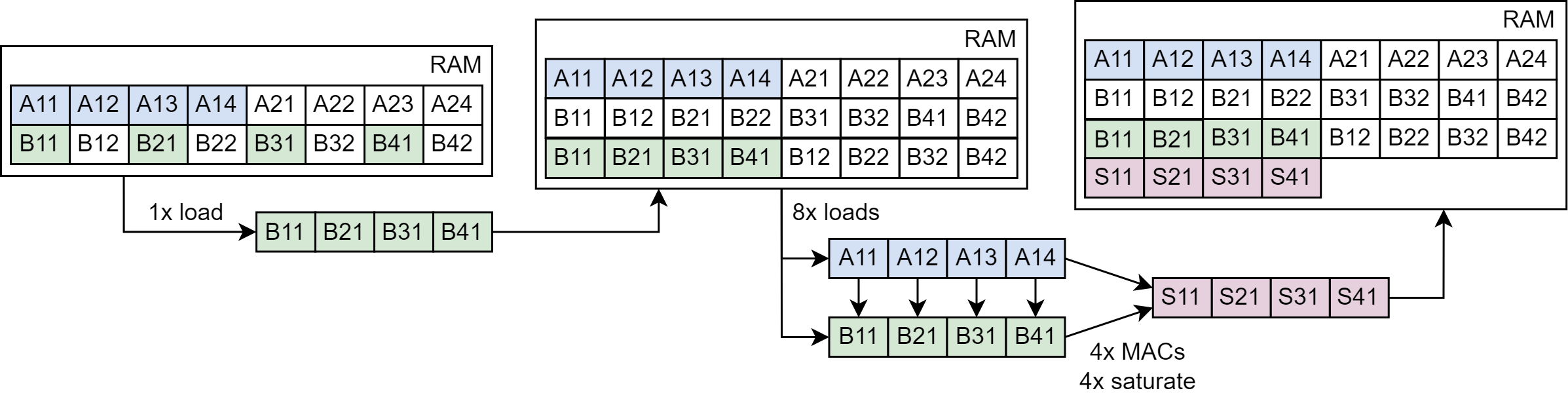}
	\caption{Matrix multiplication in Arm Cortex-M MCUs using \textit{mat\_mult\_q7\_trb}}
	\label{fig:mat_mult_q7_trb}
\end{figure}

\textit{mat\_mult\_q7\_trb} differs from \textit{arm\_mat\_mult\_q7} as it transposes the source matrix $B$ before entering in the MAC loop. As all matrices are stored in memory in height-width format, the parameters of the source matrix $B$ can not be accessed contiguously. We believe this can reduce the complexity in the calculus of the memory address of the operands during a MAC operation. The general workflow of \textit{mat\_mult\_q7\_trb} is outlined in Figure \ref{fig:mat_mult_q7_trb}.

\textit{mat\_mult\_q7\_simd} uses the SIMD feature of Armv7E-M and Armv8-M to read from memory and for MAC operations. For the signed MAC operation, the ISAs of these architectures do not feature instructions with 4x8-bit operands. However, they still feature instructions with 2x16-bit integers as operands. This means that we have to trade off the additional speedup delivered by SIMD instructions with the additional computation to perform sign extension. \textit{mat\_mult\_q7\_simd} also transposes the source matrix $B$ prior to the MAC loop; however, it also sign-extends it to 16-bit. This function is also composed of (i) one outer loop over the rows of the source matrix $A$ and (ii) an inner loop over the columns of the transposed and sign-extended matrix $B$. However, as the SMLAD instruction takes 2x16-bit operands, we applied loop unrolling on the inner loop and compute 2 MACs simultaneously. For a 4x4 sized matrix multiplication kernel, it requires 4 load operations with sign extension and 2 MACs. Algorithm \ref{alg:mat_mult_q7_simd_arm} details the $mat\_mult\_q7\_simd$.

\begin{algorithm}[h]
\small
\caption{\textit{mat\_mult\_q7\_simd(a, b, shift, r)} for Arm Cortex-M}
\label{alg:mat_mult_q7_simd_arm}
\begin{algorithmic}[1]
\STATE transpose and sign extend $b$ to 16-bit: $b\_trb = matrix\_q7\_to\_q15\_transposed(b)$
\FOR{row $i$ in $a$}
    \STATE initialize accumulator: $sum = 0$
    \STATE read and expand int-8 word from $a$ into two int-16 words: $read\_and\_pad(i, a1, a2)$
    \FOR{column $j$ in $num\_cols\_b >> 2$}
        \STATE read two int-16 from $b\_trb$: $b1 = read\_q15x2\_ia(j)$
        \STATE read two int-16 from $b\_trb$: $b2 = read\_q15x2\_ia(j)$ 
        \STATE MAC: $sum = \_\_SMLAD(a1, b1, sum)$
        \STATE MAC: $sum = \_\_SMLAD(a2, b2, sum)$
    \ENDFOR
    \FOR{column $j$ in $num\_cols\_b \% 4$}
        \STATE MAC: $sum = sum + (i * j)$
    \ENDFOR
    \STATE fix quantization format: $sum = \_\_SSAT(sum >> shift, 8)$
    \STATE copy accumulator to $r$: $^*r++ = sum$
\ENDFOR
\end{algorithmic}
\end{algorithm}

\subsubsection{\textbf{RISC-V}} The ISA of RISC-V does not support SIMD instructions. However, in contrast to Arm Cortex-M, the RV32IMCXpulp extension provides SIMD instructions for MAC operations with 4x8-bit words as operands. RISC-V SOCs featuring the RV32IMCXpulp extension are usually provided with a fabric core and a multi-core acceleration cluster tailored for ANN computation \citep{Flamand2018}. The fabric core is an MCU built around the RISC-V core and extended for efficient digital signal processing. The acceleration cluster is turned on from the fabric controller when the programmer wants to offload computation-intensive kernels. Both the fabric controller and the acceleration cluster feature the RV32IMCXpulp extension. Nevertheless, PULP-NN does not feature any built-in function for the signed and quantized matrix multiplication in int-8 format. As a consequence, we developed three additional functions: (i) \textit{mat\_mult\_q7}, (ii) \textit{mat\_mult\_q7\_trb}, and (iii) \textit{mat\_mult\_q7\_simd}. They are based on the Cortex-M implementation, but are tuned to extract the most of the acceleration cluster.

The \textit{mat\_mult\_q7} does not rely on any SIMD instruction for reading and MAC operations, nor transposes the source matrix $B$ prior to the MAC loop. It is the equivalent of the \textit{arm\_mat\_mult\_q7} but re-designed to support parallelization over the rows of the output matrix. The number of cores used in the parallelization must be a power of two ($2^n$). Similar to its Arm Cortex-M counterpart, \textit{mat\_mult\_q7} requires 8 load operations without sign extension and 4 MACs for a 4x4 sized matrix multiplication kernel.

\textit{mat\_mult\_q7\_trb} is very similar to the Arm implementation (figure \ref{fig:mat_mult_q7_trb}) - it was designed to evaluate the advantages of transposing the source matrix $B$ prior to the MAC loop. Notwithstanding, there are two main differences. Firstly, it is tuned for multi-core processing. The transpose of the source matrix $B$ is parallelized on the acceleration cluster over the row dimension and the same applies to the MAC loop. Secondly, the sign saturate is performed via a GCC built-in for the RISC-V RV32IMCXpulp (\textit{\_\_builtin\_pulp\_clip\_r}). For a 4x4 matrix multiplication kernel, this function requires 8 load operations without sign extension and 4 MACs.

When considering the \textit{mat\_mult\_q7\_simd} function (Algorithm \ref{alg:mat_mult_q7_simd_riscv}), the RISC-V RV32IMCXpulp version is more efficient as the ISA of this architecture features a SIMD instruction for the MAC of 4x8-bit operands. This function also transposes the source matrix $B$ prior to the MAC loop. Both the matrix transposing and the MAC loop is performed in the same multi-core fashion as for \textit{mat\_mult\_q7\_trb}. When compared to Arm for a 4x4 matrix multiplication kernel, it requires 2 loads without sign extension and 1 MAC operation instead of 4 loads with sign extension and 2 MACs.

\begin{algorithm}[h]
\small
\caption{\textit{mat\_mult\_q7\_simd(a, b, shift, r)} for RISC-V}
\label{alg:mat_mult_q7_simd_riscv}
\begin{algorithmic}[1]
\STATE transpose $b$: $b\_trb = transpose(b)$
\FOR{row $i$ in $a[start\_row, stop\_row]$}
    \STATE initialize accumulator: $sum = 0$
    \STATE read int-8 word from $a$: $a1 = read(i)$
    \FOR{column $j$ in $num\_cols\_b >> 2$}
        \STATE read int-8 word from $b\_trb$: $b1 = read(j)$
        \STATE MAC: $sum = \_\_builtin\_pulp\_sdotsp4(a1, b1, sum)$
    \ENDFOR
    \FOR{column $j$ in $num\_cols\_b \% 4$}
        \STATE MAC: $sum = sum + (i * j)$
    \ENDFOR
    \STATE fix quantization format: $sum = \_\_builtin\_pulp\_clip\_r(sum >> shift, 127)$
    \STATE copy accumulator to $r$: $^*r++ = sum$
\ENDFOR
\end{algorithmic}
\end{algorithm}

\subsection{Squash Activation and Vector Norm} \label{sec:squash}
The \textit{squash} activation function is designed to receive as input a 2D matrix, where each row corresponds to a vector to be squashed. For each row, the computation starts with the calculation of the vector length and ends with the squashing of each vector entry. The calculation of the vector norm uses a 32-bit integer to accumulate the sum of the power of two of each vector entry. As computing the exact value of a square root leads to severe overhead and latency in both architectures, we approximate the square root operation using the Newton-Raphson method \citep{Ram2009}.

The Newton-Raphson method is one of the most efficient techniques to approximate the zeroes of a mathematical function ($f(x)=0$) \citep{Ram2009}. It takes as input an initial approximation and then enters into an iterative process that produces successively better approximations. This iterative process is based on the evaluation of the derivative of the zero approximation. The square root calculus can be translated as finding the solution to Equation \ref{eq:newton_raph_1}, which determines Equation \ref{eq:newton_raph_2} to find the next optimal solution at each iteration of the Newton-Raphson method \citep{Ram2009}. The calculus ends whenever the absolute difference between the current zero-approximation and the zero-approximation projected for the next iteration is below 1 \citep{Ram2009}. Algorithm \ref{alg:newton_raph} details the implementation of the Newton-Raphson method.

\begin{equation}
    \label{eq:newton_raph_1}
    f(x_n)=x_n^2-n
\end{equation}

\begin{equation}
    \label{eq:newton_raph_2}
    x_{n+1} = \frac{(x_n + n)/x_n}{2}
\end{equation}

\begin{algorithm}[h]
\small
\caption{Newton-Raphson method to calculate the square root of $n$}
\label{alg:newton_raph}
\begin{algorithmic}[1]
    \STATE take $n/2$ as the initial approximate root: $x_0 = n/2$
    \STATE calculate the next approximate root according to Equation \ref{eq:newton_raph_2}: $x_1 = \frac{(x_n + n)/x_n}{2}$
    \WHILE{$x_1 < x_0$}
        \STATE update the current approximate root: $x_0 = x_1$
        \STATE update the next approximate root: $x_1 = \frac{(x_n + n)/x_n}{2}$
    \ENDWHILE
    \RETURN $x_0$
\end{algorithmic}
\end{algorithm}

As the length of a vector is calculated, each vector entry is squashed. In full-precision arithmetic, the output of \textit{squash} is in the range $[0,1]$. Therefore, the minimal precision loss is achieved if the output of \textit{squash} is quantized in absolute Q0.7 format, i.e., all bits except for the sign bit represent the fractional part of a number. To comply with this quantization format, we tuned the original \textit{squash} activation (Equation \ref{eq:squash_func}) as detailed in Equation \ref{eq:squash_quant_func}, where $i_{qn}$ and $o_{qn}$ represent the number of virtual fractional bits in the input and output data, respectively. By extending the scaling granularity to the squared norm and norm parameters, we avoid the accuracy loss when the floating-point norm is below 0. For more details on the absolute and virtual quantization formats, we refer to Section \ref{sec:quant_fw}. Equation \ref{eq:squash_quant_func} embeds the quantization of the \textit{squash} activation output in the \textit{squash} activation itself, avoiding the use of floating-point division and dynamic quantization, which introduces severe overhead. For the RISC-V implementation, the \textit{squash} kernel can be offloaded to the acceleration cluster and parallelized along the vectors of the input matrix. The number of vectors is equally divided between each core. In case a fair division is not possible, the last core handles the remaining vectors.

\begin{equation}
    v_j = \frac{||s_j|| << (o_{qn}-i_{qn})}{1 << i_{qn} + ||s_j||^{^{2}} >> i_{qn}} * s_j
    \label{eq:squash_quant_func}
\end{equation}

\subsection{Primary Capsule Layer} \label{sec:pcap_des}
A Primary capsule layer is the first capsule layer of a CapsNet. It has a convolutional kernel and can be observed as a convolutional layer with \textit{squash} activation. Nevertheless, we have to consider that a primary capsule layer is 4-dimensional, while CMSIS-NN and PULP-NN only support convolution over 3D data. This characteristic of a primary capsule layer can be observed in Fig. \ref{fig:CapsNet_Arch} - it is characterized by the (i) height, (ii) width, and (iii) depth of the capsules, as well by the (iv) number of capsules. Instead of implementing a 3D convolutional kernel to perform convolution over 4D data, we borrowed the implementation strategy of \citet{Sabour2017} and implemented primary capsule layers with 2D convolution kernels and then perform reshape. The third dimension of the data submitted to these 2D convolution kernels corresponds to the last two dimensions of the capsule layer \textit{(capsule depth x number of capsules)}. The output of the convolution is reshaped as a 2D vector where the first dimension corresponds to \textit{capsule height x capsule width x number of capsules} and the last dimension correspond to \textit{capsule depth}. These 2D vectors get squashed along the last dimension, ensuring that the magnitude of the capsule depth sums to 1. The output feature map of the squash activation is then reshaped to match the original 4D shape of the primary capsule layer. As detailed in section \ref{sec:quant_fw}, quantized primary capsule layers must scale the output feature map of the convolution prior to the squash activation to minimize the accuracy loss. For this purpose, for both Arm Cortex-M and RISC-V, our software kernel requires the programmer to pass two scaling factors: (i) one for the bias and (ii) another for the outputs.

\subsubsection{\textbf{Arm Cortex-M}}
In CMSIS-NN, there are two functions for convolutions with 8-bit data: a basic version - \textit{arm\_convolve\_HWC\_q7\_basic\_nonsquare} and another for reduced latency - \textit{arm\_convolve\_HWC\_q7\_fast\_nonsquare}. Despite the reduced latency, the later version imposes constraints on the dimension of the input and output channels, which must be multiples of 4 and 2, respectively. Considering that these restrictions are too severe and may not fit all the design scenarios, we decided to provide the API with two functions for the primary capsules: one based on the basic convolution (\textit{pcap\_q7\_basic}) and another based on the fast convolution (\textit{pcap\_q7\_fast}). These functions are already designed to leverage most of the SIMD features in Armv7E-M and Armv8-M architectures. Nevertheless, as the remaining CMSIS-NN functions, they are not ready to parallelize the convolution over multi-cores. CMSIS-NN already embeds in their convolution kernels the scaling of the output feature map. The scaling is performed as a bitwise-shift operation for the bias and the outputs. The \textit{squash} activation is detailed in section \ref{sec:squash}. For more details on the convolution kernels implemented by CMSIS-NN, we refer to the work \citep{CMSIS-NN}.

\subsubsection{\textbf{RISC-V}}
PULP-NN does not provide any function that can be directly used to implement the convolution part of a primary capsule. Convolution functions provided by PULP-NN involve the clipping of negative activations to zero, which is incompatible with the fundamentals of CapsNets. Clipping negative values limit the direction scope and distort the magnitude of the output vector, introducing an additional non-linearity that CapsNets are not designed to support. Nevertheless, RISC-V (RV32IMCXpulp) already features SIMDs and GCC builtin-functions for the dot-product of two vectors with 4x8-bit signed data. Furthermore, it also provides a builtin-function for the saturate operation. Given the background, we borrowed the PULP-NN functions implementing convolutional layers (\textit{pulp\_nn\_conv\_Co\_parallel, pulp\_nn\_conv\_Ho\_parallel, pulp\_nn\_conv\_HoWo\_parallel}) and adapted them for signed int-8 data. We updated the data type of all arguments and intermediate variables in unsigned integer to signed integer and replaced the GCC built-ins for the dot-product and saturate operations, which are now performed by \textit{\_\_builtin\_pulp\_sdotsp4} and \textit{\_\_builtin\_pulp\_clip\_r}, respectively. The batch normalization and relu activation were also updated. The new convolution functions - (i) \textit{pulp\_nn\_conv\_Co\_parallel\_q7}, (ii) \textit{pulp\_nn\_conv\_Ho\_parallel\_q7}, (iii) and \textit{pulp\_nn\_conv\_HoWo\_parallel\_q7} - maintain the same inner loop of the convolution used by PULP-NN. They are also optimized to run on the acceleration cluster. Consequently, our software kernels feature three different functions for the primary capsule layer, each one devoted to a different parallelization strategy - parallelization can be performed along the channel (\textit{pcap\_Co\_q7}), height (\textit{pcap\_Ho\_q7}), or height vs. width (\textit{pcap\_HoWo\_q7}). Pixels are equally divided by all cores. If a fair division is not possible, the last core handles the remaining. Nevertheless, primary capsule kernels can also run in the fabric controller. For more details on the inner loop of the convolution, we refer to the work \citep{PULP-NN}. The \textit{squash} activation is detailed in section \ref{sec:squash}.

\subsection{Capsule Layer} \label{sec:cap_des}
Algorithm \ref{alg:dynamic_routing} details the workflow of the kernel implementing a capsule layer (\textit{capsule\_layer\_q7}). This workflow is mainly implemented by a series of four support functions, starting with the \textit{calc\_inputs\_hat} which calculates the prediction vectors, which is then submitted to the dynamic routing. The dynamic routing starts with the \textit{calc\_coupling\_coefs} which computes the coupling coefficients by applying a softmax function to the logits representing the agreement between capsules in consecutive layers. These coupling coefficients and the prediction vectors are further used as inputs to \textit{calc\_caps\_output}, which returns the output vectors of the capsule for the current routing iteration. If the dynamic routing is not in the last iteration, the function \textit{calc\_agreement\_w\_prev\_caps} calculates the agreement between capsules in layer $L+1$ and capsules in layer $L$. Each one of these support functions is detailed below.

\begin{algorithm}[h]
\small
\caption{\textit{capsule\_layer\_q7(args)} for Arm Cortex-M and RISC-V}
\label{alg:dynamic_routing}
\begin{algorithmic}[1]
    \STATE initialize logits $b_{ij}$ as zeroes: $memset(b, 0, num\_caps * input\_num\_caps)$
    \STATE calculate prediction vectors: $calc\_inputs\_hat(b, layer\_input, shift, pred\_vectors)$
    \FOR{routing $r$ in $num\_routings$}
        \STATE calculate coupling coefficients $c\_{ij}$: $calc\_coupling\_coefs(b, c)$
        \STATE calculate output vectors: $calc\_caps\_output(pred\_vectors, c, shift, out\_vectors)$
        \IF{$r < num\_routings - 1$}
            \STATE calculate agreement between capsules: $calc\_agreement\_w\_prev\_caps(pred\_vectors, out\_vectors, shift, b)$
        \ENDIF
    \ENDFOR
\end{algorithmic}
\end{algorithm}

\subsubsection{\textbf{calc\_inputs\_hat}} Multiplies the input of the capsule layer by the respective weight matrix. These weights connect each feature of each capsule in layer $L$ to each feature of each capsule in layer $L+1$. As a consequence, the original weight matrix is 4D \textit{[total capsules in layer $L+1$, total capsules in layer $L$, capsule size in layer $L+1$, capsule size in layer $L$]} and not directly supported by our matrix multiplication kernels. Nevertheless, performing a direct 4D matrix multiplication would require the replication of the input feature map for each capsule, increasing the memory footprint. Instead, \textit{calc\_inputs\_hat} regards the 2 inner dimensions of the weight matrix as valid matrix multiplication dimensions and the outer dimensions as batch size. \textit{calc\_inputs\_hat} has an outer loop that iterates over all capsules of layer $L+1$ and an inner loop that iterates over all capsules of layer $L$. In the inner loop, the weights connecting a capsule $i$ in layer $L$ to a capsule $j$ in layer $L+1$ are multiplied by the input features of capsule $j$ (output features of capsule $i$). The multiplication of these two matrices uses the fastest of the kernels described in section \ref{sec:mat_mul} - \textit{mat\_mult\_q7\_trb} for Arm and \textit{mat\_mult\_q7\_simd} for RISC-V. The resulting prediction vector is then submitted to the iterative dynamic routing where the output vector will be iteratively updated. 

\subsubsection{\textbf{calc\_coupling\_coefs}} First function of the the dynamic routing and computes the coupling coefficients $c_{ij}$. More specifically, \textit{calc\_coupling\_coefs} applies a softmax function over the logits $b_{ij}$ that measure the agreement between capsules at different layers. Each logit $b_{ij}$ measures the agreement between a capsule $i$ in the previous layer (layer $L$) and a capsule $j$ of the current layer (layer $L+1$). Applying a softmax function to these logits ensures that the importance of a capsule $i$ in layer $L$ for all capsules of the current layer $L+1$ sums to 1. For the Armv7E-M and Armv8-M, we make direct use of the \textit{arm\_softmax\_q7} function provided by CMSIS-NN. In contrast, PULP-NN does not provide any function implementing softmax. We developed a \textit{softmax} function based on the Arm implementation.

\subsubsection{\textbf{calc\_caps\_output}} Multiplies the coupling coefficients with the prediction vectors, returning the output vectors. If coupling coefficients are already expressed in a 2D matrix, the same is not applied to the prediction vector, which is represented by a 3D matrix \textit{[total capsules in layer $L+1$, total capsules in layer $L$, capsule size in layer $L+1$}]. Consequently, \textit{calc\_caps\_output} regards the 2 inner dimensions as the valid matrix multiplication dimensions and the outer dimension as batch size. Matrix multiplication is performed using the fastest of the kernels described in section \ref{sec:mat_mul}. When the matrix multiplication loop finishes, the output vectors are squashed using the kernel described in section \ref{sec:squash}.

\subsubsection{\textbf{calc\_agreement\_w\_prev\_caps}} Determines the degree of agreement between the weighted output of a capsule in layer $L$ (prediction vectors) with the output of capsules (output vectors) in layer $L+1$. The agreement is determined as the product of the respective matrices. As the prediction vector is represented by a 3D matrix \textit{[total capsules in layer $L+1$, total capsules in layer $L$, capsule size in layer $L+1$}], \textit{calc\_caps\_output} consider the inner two dimensions as the valid matrix multiplication dimensions and the outer dimension as batch size. The result is then summed up to the logits $b_{ij}$. For this purpose, we rely on 2D matrix addition kernels. 

\section{CapsNet Quantization} \label{sec:quant_fw}
The quantization framework is detailed in Algorithm \ref{alg:quant_fw}. It takes as input a CapsNet developed in TensorFlow and a quantization dataset and returns the quantized weights and bias, as well as the output and bias shifts. Except for primary capsule and capsule layers, this framework considers activation functions as independent layers. Embedding an activation in a dense or convolutional layer may return a faulty quantized model.

\begin{algorithm}[b]
\small
\caption{Quantization framework}
\label{alg:quant_fw}
\begin{algorithmic}[1]
    \STATE load CapsNet from .h5 file
    \STATE load quantization reference dataset
    \STATE quantize weights: $wq_{ij} = round(w_{ij} / S)$
    \STATE quantize bias: $bq_{ij} = round(b_{ij} / S)$
    \STATE calculate amount of output/bias shift:
    \begin{ALC@g}
        \FOR{layer $l$ in CapsNet}
            \FOR{matrix multiplication, matrix addition or convolution $k$}
                \STATE get Qm.n format of the input, output, and intermediates
                \STATE calculate amount of output shift: $out\_s = f_{ia} + f_{ib} - f_o$ \label{eq:out_shift}
                \STATE calculate amount of bias shift: $bias\_s = f_{ia} + f_{ib} - f_b$ \label{eq:bias_shift}
            \ENDFOR
        \ENDFOR
    \end{ALC@g}
    \RETURN $wq, bq, s$
\end{algorithmic}
\end{algorithm}

As detailed in Algorithm \ref{alg:quant_fw}, after loading the model and the dataset, the quantization process starts with the quantization of the model's weights. The quantization itself is performed as detailed in Algorithm \ref{alg:quant}. For each layer, the framework searches for the maximum absolute weight ($max\_abs$) and calculates the number $m$ of integer bits required to represent it. As we are using a power-of-two scaling ($A x 2^n$), the number of integer bits to represent a value $A$ is calculated as the ceiling of $log_2(max_{abs})$. The remaining $n$ bits are considered for the fractional part. Nevertheless, to increase precision in layers with very small weights (maximum absolute weight lower than $1/127$), our framework virtually increases the number of fractional bits, bringing the maximum norm of the quantized weights closer to the saturation barrier without never going over it. In absolute or physical terms, every weight is still in $Q0.7$ format as it still fits in eight bits, but virtually the quantization format can surpass the eight-bit barrier. It must be noted that the first bit of the eight available is used for the sign. This $Qm.n$ quantization format allows the representation of values comprised in the uniform and symmetric range $[-max\_abs, max\_abs]$. After getting the quantization $Qm.n$ format, the framework proceeds to the calculus of the quantized weights, where the floating-point number $A$ is multiplied by 2 powered to the number of fractional bits $n$ ($A x 2^n$). The quantization finishes with the clipping of values to the range $[-128, 127]$. The quantization of the model's bias is performed under the same workflow.

\begin{algorithm}[h]
\small
\caption{Quantization algorithm}
\label{alg:quant}
\begin{algorithmic}[1]
    \STATE get Qm.n format:
    \begin{ALC@g}
        \STATE find maximum absolute value $max_{abs}$
        \STATE find amount $m$ of bits to represent the range $[-max_{abs}, max_{abs}]$: the ceiling of $log_2(max_{abs})$
        \STATE calculate amount $n$ of bits for the fractional part:
        \begin{ALC@g}
            \STATE $n = 7 - m$
            \WHILE{($max_{abs} >> (n+1)) <= 127$}
                \STATE $n++$
            \ENDWHILE
        \end{ALC@g}
    \end{ALC@g}
    \STATE convert to quantized integer:
    \begin{ALC@g}
        \STATE multiply floating-point values $A$ by 2 powered to the number of fractional bits: $A x 2^n$
        \STATE clip quantized integers for the range [-128, 127]
    \end{ALC@g}
\end{algorithmic}
\end{algorithm}

This quantization strategy enables the addition and multiplication of quantized values with minimal overhead. However, the results must be scaled to avoid saturation problems. As we are using a power-of-two-scaling, scaling can be performed as a bitwise shift operation. In the context of CapsNets, scaling takes part of the convolutional, primary capsule, and capsule layers as these layers apply matrix multiplication and/or addition. Convolutional and primary capsule layers only go through a single matrix multiplication process, and therefore only take one scaling factor for the bias and another for the output. However, capsule layers involve at least three distinctive matrix multiplications and one matrix addition. When considering the four support functions of a capsule layer kernel, only \textit{calc\_coupling\_coefs} does not require any scaling factor. \textit{calc\_inputs\_hat} requires one output scaling factor and  \textit{calc\_caps\_output} requires one for each iteration of the dynamic routing. As \textit{calc\_agreement\_w\_prev\_caps} comprises matrix multiplication and addition, it requires two output scaling factors per iteration of the dynamic routing unless for the last one.

The scaling factors are calculated as detailed in Algorithm \ref{alg:quant_fw}. For each matrix multiplication or addition in each layer of the network, the reference quantization dataset is used to determine the maximum absolute value for the input and output data. Then, it applies Algorithm \ref{alg:quant} to get the quantization format ($Qm.n$) for this data. At this step, the framework is ready to calculate the number of right shifts for every matrix multiplication/addition. Considering $f_{ia}$ as the number of fractional bits for the input matrix $A$, $f_{ib}$ as the number of fractional bits for the input matrix $B$, and $f_o$ as the number of fractional bits for the output, the amount of right shifts (output scaling factor) is calculated as detailed in line \ref{eq:out_shift} of Algorithm \ref{alg:quant_fw}. The same process is carried out to calculate the left bias shift of convolutional and primary capsule layers. The bias shift is calculated as detailed in line \ref{eq:bias_shift} of Algorithm \ref{alg:quant_fw}. $f_b$ represents the number of fractional bits for the bias.

\section{Evaluation} \label{sec:results}
We evaluated the software kernels and the quantization framework on three imaging datasets: (i) MNIST \citep{mnist}, (ii) smallNORB, and \citep{smallnorb} (iii) CIFAR-10 \citep{cifar10}. MNIST is a large database of handwritten digits between 0 and 9. It contains 60k training images and 10k testing images, both shaped as $28x28x1$. smallNORB is a dataset designed to develop applications for 3D object recognition from shape. It contains 5 generic categories of toys - (i) four-legged animals, (ii) human figures, (iii) airplanes, (iv) trucks, and (v) cars - imaged under varying light conditions, elevations, and azimuths. Both training and testing batches are composed of 24.3k images, shaped as $96x96x2$. CIFAR-10 consists of 60000 $32x32x3$ colour images, representing 10 different objects. There are 50k images for training and 10k for testing.

For each dataset, we trained a full-precision and baseline reference CapsNet. We do not intend to build CapsNets with state-of-the-art accuracy on these datasets but rather build a CapsNet that after quantization fits in the limited memory resources of the target low-power MCUs without compromising the accuracy. We have built CapsNets that after quantization take no more than 80\% of the RAM available in the least performing target MCU. At the moment of this evaluation, our software kernels do not support tiling. Thus, we have to ensure that both the CapsNet parameters and at least one sampling image can fit in the available RAM without compromising the execution of our software kernels. Table \ref{tab:capsnet_eval} details the architecture and hyper-parameters of the CapsNets used for testing.

\begin{table}[h]
\caption{Baseline reference CapsNets for evaluation}
\label{tab:capsnet_eval}
\resizebox{\linewidth}{!}{
\begin{tabular}{@{}lclclclll@{}}
\toprule
\textbf{Dataset}              & \multicolumn{2}{l}{\textbf{Convolutional Layers}}                                                                                                                                                & \multicolumn{2}{l}{\textbf{Primary Capsule Layers}}                                                                                      & \multicolumn{2}{l}{\textbf{Capsule Layers}}                                                                               & \multicolumn{2}{l}{\textbf{Optimizer}} \\ \cmidrule(l){2-9} 
\multicolumn{1}{c}{\textbf{}} & \multicolumn{1}{l}{\textbf{Layer \#}} & \textbf{Configuration}                                                                                                                                   & \multicolumn{1}{l}{\textbf{Layer \#}} & \textbf{Configuration}                                                                           & \multicolumn{1}{l}{\textbf{Layer \#}} & \textbf{Configuration}                                                            & \textbf{Type}       & \textbf{LR}      \\ \midrule
\textbf{MNIST}                & 1                                     & \begin{tabular}[c]{@{}l@{}}Filters: 16\\ Kernel Size: 7\\ Stride: 1\\ Activation: ReLU\end{tabular}                                                      & 2                                     & \begin{tabular}[c]{@{}l@{}}Capsules: 16\\ Dimension: 4\\ Kernel Size: 7\\ Stride: 2\end{tabular} & 3                                     & \begin{tabular}[c]{@{}l@{}}Capsules: 10\\ Dimension: 6\\ Routings: 3\end{tabular} & Adam                & 0.001            \\ \midrule
\textbf{smallNORB}            & 1                                     & \begin{tabular}[c]{@{}l@{}}Filters: 32\\ Kernel Size: 7\\ Stride: 1\\ Activation: ReLU\end{tabular}                                                      & 2                                     & \begin{tabular}[c]{@{}l@{}}Capsules: 16\\ Dimension: 4\\ Kernel Size: 7\\ Stride: 2\end{tabular} & 3                                     & \begin{tabular}[c]{@{}l@{}}Capsules: 5\\ Dimension: 6\\ Routings: 3\end{tabular}  & Adam                & 0.00025          \\ \midrule
\textbf{CIFAR-10}             & 1-4                                   & \begin{tabular}[c]{@{}l@{}}Filters: {[}32, 32, 64, 64{]}\\ Kernel Size: 4 x {[}3{]}\\ Stride: {[}1, 1, 2, 2{]}\\ Activation: 4 x {[}ReLU{]}\end{tabular} & 5                                     & \begin{tabular}[c]{@{}l@{}}Capsules: 16\\ Dimension: 4\\ Kernel Size: 3\\ Stride: 2\end{tabular} & 6                                     & \begin{tabular}[c]{@{}l@{}}Capsules: 10\\ Dimension: 5\\ Routings: 3\end{tabular} & Adam                & 0.00025            \\ \bottomrule
\end{tabular}
}
\end{table}

The evaluation included three Arm Cortex-M MCUs: STM32L4R5ZIT6U (Cortex-M4 @ 120 MHz, 640 KB RAM), STM32H755ZIT6U (Cortex-M7 @ 480 MHz, 1 MB RAM), and STM32L552ZET6QU (Cortex-M33 @ 110 MHz, 512 KB RAM). Hence, our evaluation encompasses CPUs featuring Armv7-M architecture (Cortex-M4 and Cortex-M7), and the more recent Armv8-M (Cortex-M33). At the time of this evaluation, the market's scarcity of in-silicon MCUs featuring the RISC-V RV32IMCXpulp ISA severely limited the testing setup. For RISC-V, our software kernels were tested on a single development board featuring a GAP-8 MCU (GreenWaves GAPuino v1), with a fabric controller running at 250 MHz and an acceleration cluster with 8 cores running at 170 MHz. All cores share 512 KB of RAM. Although there are several open-source RISC-V cores using RISC-V RV32IMC ISA, using soft-cores deployed in an FPGA to test our software kernels would not enable a fair comparison with the Arm Cortex-M MCUs.

\subsection{Quantization}
The quantization framework was evaluated in terms of memory footprint reduction and accuracy loss. As detailed in Section \ref{sec:quant_fw}, the quantization mechanism returns a series of output and bias shifts that must be performed during inference time. We consider these parameters as part of the memory footprint inherent to the quantized CapsNet. Table \ref{tab:quant_fw_res} details the accuracy and memory footprint of the full precision CapsNets against their quantized counterparts. While a full-precision CapsNet requires 4 bytes per parameter, the quantized version only requires 1 byte. This represents a theoretical memory saving of 75\%, which is very near to the memory saving achieved by our quantization framework (74.99\%). This demonstrates that the size of the additional parameters required by quantized CapsNets for bias and output shifting is almost negligible. This applies to conventional 3-layered CapsNets but also deeper CapsNets as the one developed for the CIFAR-10 dataset.

In terms of accuracy, the maximum loss was achieved for the MNIST dataset (0.18\%) and the minimum for the smallNORB (0.07\%). This shows that the accuracy loss resulting from the quantization process does not necessarily grow with the complexity of the dataset. Nevertheless, it is visible that CapsNets do not perform so-well on complex datasets like CIFAR-10. CapsNets like to account for everything in the image so it does better when there is no background noise or the background does not constantly change. As this is not the case with CIFAR-10, CapsNets will perform poorly when compared to other less noisy datasets.

\begin{table}[h]
\centering
\caption{Evaluation of the quantization framework}
\label{tab:quant_fw_res}
\resizebox{0.70\linewidth}{!}{
\begin{tabular}{@{}lcccccc@{}}
\toprule
\textbf{Dataset}   & \multicolumn{3}{c}{\textbf{Memory Footprint (KB)}}         & \multicolumn{3}{c}{\textbf{Accuracy}}                    \\ \cmidrule(l){2-7} 
                   & \textbf{Float-32} & \textbf{Int-8} & \textbf{Int-8 Saving} & \textbf{Float-32} & \textbf{Int-8} & \textbf{Int-8 Loss} \\ \midrule
\textbf{MNIST}     & 1187.20           & 296.82         & 74.99\%               & 99.01\%           & 98.83\%        & 0.18\%              \\ \midrule
\textbf{smallNORB} & 1182.34           & 295.61         & 74.99\%               & 92.56\%           & 92.49\%        & 0.07\%              \\ \midrule
\textbf{CIFAR-10}  & 461.19            & 115.33         & 74.99\%               & 78.54\%           & 78.38\%        & 0.16\%             \\ \bottomrule
\end{tabular}
}
\end{table}

\subsection{Software Kernels}
\subsubsection{\textbf{Matrix Multiplication}} \label{sec:mat_mul_results}
To select the fastest matrix multiplication algorithm for each target architecture, we considered the multiplication of two randomized matrices with dimensions 20x30 and 30x40. The content of each matrix was kept constant during the entire test and the latency was measured in milliseconds and clock cycles. As detailed in Section \ref{sec:MM_ARM}, for the Arm Cortex-M architecture we tested three different algorithms: (i) \textit{mat\_mult\_q7\_simd}, (ii) \textit{mat\_mult\_q7\_trb}, and (iii) \textit{arm\_mat\_mult\_q7}. The last one is used as a baseline reference as it does not apply any SIMD or matrix transposing operation to simplify memory access during MAC. Results are detailed in Table \ref{tab:mat_mul_arm_res}.

\begin{table}[h]
\caption{Evaluation of the matrix multiplication functions for Arm Cortex-M MCUs}
\label{tab:mat_mul_arm_res}
\resizebox{\linewidth}{!}{%
\begin{tabular}{@{}lcccccc@{}}
\toprule
              & \multicolumn{2}{c}{\textbf{STM32L4R5ZIT6U}}                                           & \multicolumn{2}{c}{\textbf{STM32H755ZIT6U}}                                           & \multicolumn{2}{c}{\textbf{STM32L552ZET6QU}}                                          \\
                             & \multicolumn{2}{c}{\textit{(Armv7E-M, Cortex-M4)}}                                    & \multicolumn{2}{c}{\textit{(Armv7E-M, Cortex-M7)}}                                    & \multicolumn{2}{c}{\textit{(Armv8-M, Cortex-M33)}}                                    \\ \cmidrule(l){2-7} 
                             & \multicolumn{1}{l}{\textbf{Clock Cycles}} & \multicolumn{1}{l}{\textbf{Milliseconds}} & \multicolumn{1}{l}{\textbf{Clock Cycles}} & \multicolumn{1}{l}{\textbf{Milliseconds}} & \multicolumn{1}{l}{\textbf{Clock Cycles}} & \multicolumn{1}{l}{\textbf{Milliseconds}} \\ \midrule
\textbf{arm\_mat\_mult\_q7}  & 704395                                    & 5.87                                      & 790989                                    & 1.65                                      & 654738                                    & 5.96                                      \\ \midrule
\textbf{mat\_mult\_q7\_trb}  & 655415                                    & 5.47                                      & 574532                                    & 1.20                                      & 605769                                    & 5.51                                      \\ \midrule
\textbf{mat\_mult\_q7\_simd} & 730562                                    & 6.09                                      & 757482                                    & 1.58                                      & 697749                                    & 6.35                                      \\ \bottomrule
\end{tabular}
}
\end{table}

Contrary to what was expected, for the Arm Cortex-M implementation, the maximum performance is not observed for the algorithm relying on SIMD instructions. This occurs because the Arm ISA only supports MAC instructions with 16-bit operands, requiring the signal extension of each 8-bit operand. Despite providing SIMD instructions for the sign extension operation, it still introduces too much performance overhead. \textit{mat\_mult\_q7\_simd} is even slower than the \textit{arm\_mat\_mult\_q7}, which does not feature any optimization. The \textit{mat\_mult\_q7\_trb} and \textit{arm\_mat\_mult\_q7} functions are, on average, 1.15x and 1.04x faster than this algorithm, respectively. The maximum performance is registered for the algorithm not based on SIMD instructions but transposes the second input matrix before the matrix multiplication process. By simplifying the calculus of memory addresses during MAC, \textit{mat\_mult\_q7\_trb} is on average 1.15x faster than the approach based on SIMD instructions, and 1.10x faster than the Arm CMSIS implementation.

Regarding the RISC-V implementation, the results are detailed in Table \ref{tab:mat_mul_risc_res}. As GAP-8 SoC features an acceleration cluster tailored for neural network computation, we tested our matrix multiplication functions leveraging the cluster. Tests were performed in single-core and octa-core settings. Opposed to Arm Cortex-M, the fastest algorithm here is based on SIMD instructions - \textit{mat\_mult\_q7\_simd}. This happens because the RISC-V RV32IMCXpulp features SIMD instructions for the simultaneous MAC of 8-bit operands, in contrast to Arm Cortex-M that only supports 16-bit operands and requires data expansion. For the octa-core setting, \textit{mat\_mult\_q7\_simd} is $2.10x$ faster than \textit{mat\_mult\_q7\_trb} and $2.05x$ faster than \textit{mat\_mult\_q7}. In a single-core configuration, the speedup of \textit{mat\_mult\_q7\_simd} in relation to the other implementations is slightly higher: $2.20x$ faster than \textit{mat\_mult\_q7\_trb} and $2.15x$ faster than \textit{mat\_mult\_q7}. When comparing the performance of the single-core with the octa-core implementation, we notice that the octa-core is 6.32x to 6.63x faster than the single-core. The combination of the RISC-V RV32IMCXpulp SIMD instructions with multi-core processing can outperform the STM32H755ZIT6U even though this Arm MCU runs at a much higher clock frequency than the RISC-V MCU (480 MHz vs. 170 MHz).

\begin{table}[h]
\centering
\caption{Evaluation of the matrix multiplication functions for RISC-V MCUs}
\label{tab:mat_mul_risc_res}
\resizebox{0.75\linewidth}{!}{%
\begin{tabular}{@{}lcccc@{}}
\toprule
\textbf{}                    & \multicolumn{2}{c}{\textbf{GAP-8 (single-core)}}                                      & \multicolumn{2}{c}{\textbf{GAP-8 (octa-core)}}                                        \\
                             & \multicolumn{2}{c}{\textit{RISC-V RV32IMCXpulp}}                                      & \multicolumn{2}{c}{\textit{RISC-V RV32IMCXpulp}}                                      \\ \cmidrule(l){2-5} 
                             & \multicolumn{1}{l}{\textbf{Clock Cycles}} & \multicolumn{1}{l}{\textbf{Milliseconds}} & \multicolumn{1}{l}{\textbf{Clock Cycles}} & \multicolumn{1}{l}{\textbf{Milliseconds}} \\ \midrule
\textbf{mat\_mult\_q7}       & 696951                                    & 4.10                                      & 105250                                    & 0.62                                      \\ \midrule
\textbf{mat\_mult\_q7\_trb}  & 715602                                    & 4.21                                      & 107784                                    & 0.64                                      \\ \midrule
\textbf{mat\_mult\_q7\_simd} & 323844                                    & 1.91                                      & 51238                                     & 0.31                                      \\ \bottomrule
\end{tabular}
}
\end{table}

\subsubsection{\textbf{Primary Capsule Layer}}
We tested the primary capsule kernels in terms of latency on the three previously described CapsNets. For the Armv7E-M and Armv8-M architectures we developed two different functions to implement a primary capsule layer: (i) \textit{pcap\_q7\_basic} and (ii) \textit{pcap\_q7\_fast}. The former is a slower version without constraints on the input and output channel sizes. The latter is a faster version but only supports multiples of 4 and 2 for the input and output channel - a constraint that is met by the three CapsNets used for evaluation. Table \ref{tab:pcap_arm_res} details the latency for the three Arm Cortex-M MCUs in the classification of a single image. 

As expected, \textit{pcap\_q7\_fast} outperforms the baseline reference for all datasets - it is at least 1.08x faster. Nevertheless, the most noticeable impact on the latency is caused by the size of the kernel. The biggest kernel, inherent to the smallNORB CapsNet, is 2.73x bigger than the smallest of the kernels, designed for the CIFAR-10 dataset. However, the computation of the smallNORB kernel is 33.32x to 34.23x slower than the computation of the CIFAR-10 kernel. This means that the computation time does not grow linear with the kernel size of the primary capsule layer, but rather exponentially. Given the background, big kernels must be avoided whenever possible.

\begin{table}[t]
\caption{Evaluation of the primary capsule functions for Arm Cortex-M MCUs}
\label{tab:pcap_arm_res}
\resizebox{\linewidth}{!}{
\begin{tabular}{@{}llcccccc@{}}
\toprule
\textbf{}              & \textbf{}                & \multicolumn{2}{c}{\textbf{STM32L4R5ZIT6U}}                                           & \multicolumn{2}{c}{\textbf{STM32H755ZIT6U}}                                           & \multicolumn{2}{c}{\textbf{STM32L552ZET6QU}}                                          \\
\textbf{Dataset}       &                          & \multicolumn{2}{c}{\textit{(Armv7E-M, Cortex-M4)}}                                    & \multicolumn{2}{c}{\textit{(Armv7E-M, Cortex-M7)}}                                    & \multicolumn{2}{c}{\textit{(Armv8-M, Cortex-M33)}}                                    \\ \cmidrule(l){3-8} 
                       &                          & \multicolumn{1}{l}{\textbf{Clock Cycles}} & \multicolumn{1}{l}{\textbf{Milliseconds}} & \multicolumn{1}{l}{\textbf{Clock Cycles}} & \multicolumn{1}{l}{\textbf{Milliseconds}} & \multicolumn{1}{l}{\textbf{Clock Cycles}} & \multicolumn{1}{l}{\textbf{Milliseconds}} \\ \midrule
\textbf{MNIST}         & \textit{pcap\_q7\_basic} & 65.79M                                    & 548.25                                    & 63.49M                                    & 132.29                                    & 51.34M                                    & 466.77                                    \\
\textit{7x7x16x64 (M)} & \textit{pcap\_q7\_fast}  & 60.12M                                    & 500.97                                    & 57.57M                                    & 119.94                                    & 46.65M                                    & 424.13                                    \\ \midrule
\textbf{smallNORB}     & \textit{pcap\_q7\_basic} & 406.35M                                   & 3386.29                                   & 389.62M                                   & 811.70                                    & 316.95M                                   & 2881.32                                   \\
\textit{7x7x32x64 (L)} & \textit{pcap\_q7\_fast}  & 372.55M                                   & 3104.57                                   & 355.22M                                   & 740.03                                    & 289.06M                                   & 2627.78                                   \\ \midrule
\textbf{CIFAR-10}      & \textit{pcap\_q7\_basic} & 12.09M                                    & 100.75                                    & 11.40M                                    & 23.75                                     & 9.26M                                     & 84.17                                     \\
\textit{3x3x64x64 (S)} & \textit{pcap\_q7\_fast}  & 11.18M                                    & 93.19                                     & 10.50M                                    & 21.87                                     & 8.50M                                     & 77.30                                     \\ \bottomrule
\end{tabular}
}
\end{table}

For the RISC-V RV32IMCXpulp architecture, we developed three different functions implementing a primary capsule layer. As detailed in Section \ref{sec:pcap_des}, each of them parallelizes a different spatial dimension of the output feature map over the cores of the acceleration cluster. Table \ref{tab:pcap_risc_res} details the latency of the three implemented functions on the three reference CapsNets. As expected, it is not possible to mention a function that is the fastest for every dataset or CapsNet architecture. As the latency is directly impacted by the spatial dimension that is parallelized, it all depends on the shape of the input feature map. \textit{pcap\_ho\_q7} is simultaneously the fastest function for the MNIST dataset and the slowest for the other two. Similar to Arm, the latency grows exponentially with the kernel size of the primary capsule layer. While the kernel of the smallNORB is only 2.73x bigger than the kernel of the CIFAR-10, the computation of the smallNORB kernel is at least 33.39x slower (single-core). This effect is not attenuated in a multi-core configuration. 

\begin{table}[h]
\centering
\caption{Evaluation of the primary capsule functions for RISC-V MCUs}
\label{tab:pcap_risc_res}
\resizebox{0.8\linewidth}{!}{
\begin{tabular}{@{}llcccc@{}}
\toprule
\textbf{}              & \textbf{}               & \multicolumn{2}{c}{\textbf{GAP-8 (single-core)}}                                      & \multicolumn{2}{c}{\textbf{GAP-8 (octa-core)}}                                        \\
\textbf{Dataset}       &                         & \multicolumn{2}{c}{\textit{RISC-V RV32IMCXpulp}}                                      & \multicolumn{2}{c}{\textit{RISC-V RV32IMCXpulp}}                                      \\ \cmidrule(l){3-6} 
                       &                         & \multicolumn{1}{l}{\textbf{Clock Cycles}} & \multicolumn{1}{l}{\textbf{Milliseconds}} & \multicolumn{1}{l}{\textbf{Clock Cycles}} & \multicolumn{1}{l}{\textbf{Milliseconds}} \\ \midrule
\textbf{MNIST}         & \textit{pcap\_co\_q7}   & 9.45M                                     & 55.59                                     & 1.58M                                     & 9.27                                      \\
\textit{7x7x16x64 (M)} & \textit{pcap\_ho\_q7}   & 9.40M                                     & 55.27                                     & 1.19M                                     & 7.02                                      \\
                       & \textit{pcap\_howo\_q7} & 9.49M                                     & 55.85                                     & 1.18M                                     & 6.95                                      \\ \midrule
\textbf{smallNORB}     & \textit{pcap\_co\_q7}   & 57.69M                                    & 339.35                                    & 9.40M                                     & 55.32                                     \\
\textit{7x7x32x64 (L)} & \textit{pcap\_ho\_q7}   & 58.27M                                    & 342.76                                    & 11.48M                                    & 67.53                                     \\
                       & \textit{pcap\_howo\_q7} & 57.70M                                    & 339.39                                    & 11.40M                                    & 67.07                                     \\ \midrule
\textbf{CIFAR-10}      & \textit{pcap\_co\_q7}   & 1.73M                                     & 10.15                                     & 0.27M                                     & 1.59                                      \\
\textit{3x3x64x64 (S)} & \textit{pcap\_ho\_q7}   & 1.74M                                     & 10.26                                     & 0.43M                                     & 2.55                                      \\
                       & \textit{pcap\_howo\_q7} & 1.72M                                     & 10.15                                     & 0.22M                                     & 1.30                                      \\ \bottomrule
\end{tabular}
}
\end{table}

When compared to Arm, the RISC-V implementation completely outperform by almost two orders of magnitude. As the RISC-V RV32IMCXpulp ISA provides SIMD instructions specifically tailored for the simultaneous MAC of two 4x8-bit operands, it requires far less time for the same matrix multiplication kernel. For a single-core setting on the smallNORB dataset, STM32L552ZET6QU is 5.01x slower than the RISC-V MCU (\textit{pcap\_q7\_fast vs. pcap\_co\_q7}).

\subsection{Capsule Layer}
We tested the function implementing a capsule layer on the three reference datasets. Tables \ref{tab:cap_arm_res} and \ref{tab:cap_risc_res} detail the results for the Arm Cortex-M and RISC-V MCUs, respectively. Considering only the single-core scenario, we observe that the RISC-V implementation outperforms Arm. This arises as a direct consequence of the more efficient matrix multiplication kernel deployed in RISC-V (Section \ref{sec:mat_mul_results}). As detailed in Section \ref{sec:cap_des}, three of the four support functions composing the capsule layer rely on matrix multiplication. As matrix multiplication is more efficient in RISC-V MCUs, due to the use of appropriate SIMD, the capsule layer function will deliver results with lower latency. When comparing the fastest of the Arm Cortex-M MCUs in terms of clock cycles (STM32L552ZET6QU) against the RISC-V MCU in single-core mode, the latter is on average 3.95x faster. Regarding the octa-core setting of the RISC-V MCU, we can observe that this setting is on average 7.43x faster than the single-core setting. This means that the \textit{cap\_parallel\_q7} function is very near to the theoretical perfect parallelization over all the cores of the acceleration cluster.

\begin{table}[h]
\caption{Evaluation of the capsule functions for Arm Cortex-M MCUs}
\label{tab:cap_arm_res}
\resizebox{\linewidth}{!}{
\begin{tabular}{@{}llclclcl@{}}
\toprule
\textbf{}                & \textbf{}        & \multicolumn{2}{c}{\textbf{STM32L4R5ZIT6U}}                            & \multicolumn{2}{c}{\textbf{STM32H755ZIT6U}}                            & \multicolumn{2}{c}{\textbf{STM32L552ZET6QU}}                           \\
\textbf{Dataset}         &                  & \multicolumn{2}{c}{\textit{(Armv7E-M, Cortex-M4)}}                     & \multicolumn{2}{c}{\textit{(Armv7E-M, Cortex-M7)}}                     & \multicolumn{2}{c}{\textit{(Armv8-M, Cortex-M33)}}                     \\ \cmidrule(l){3-8} 
                         &                  & \multicolumn{1}{l}{\textbf{Clock Cycles}} & \textbf{Milliseconds}      & \multicolumn{1}{l}{\textbf{Clock Cycles}} & \textbf{Milliseconds}      & \multicolumn{1}{l}{\textbf{Clock Cycles}} & \textbf{Milliseconds}      \\ \midrule
\textbf{MNIST}           & \textit{cap\_q7} & 40.63M                                    & \multicolumn{1}{c}{338.56} & 49.63M                                    & \multicolumn{1}{c}{103.40} & 23.54                                     & \multicolumn{1}{c}{213.97} \\
\textit{10x1024x6x4 (L)} &                  & \multicolumn{1}{l}{}                      &                            & \multicolumn{1}{l}{}                      &                            & \multicolumn{1}{l}{}                      &                            \\ \midrule
\textbf{smallNORB}       & \textit{cap\_q7} & 32.12M                                    & \multicolumn{1}{c}{267.65} & 43.49M                                    & \multicolumn{1}{c}{90.60}  & 20.45                                     & \multicolumn{1}{c}{185.90} \\
\textit{5x1600x6x4 (M)}  &                  & \multicolumn{1}{l}{}                      &                            & \multicolumn{1}{l}{}                      &                            & \multicolumn{1}{l}{}                      &                            \\ \midrule
\textbf{CIFAR-10}        & \textit{cap\_q7} & 9.55M                                     & \multicolumn{1}{c}{79.58}  & 14.22M                                    & \multicolumn{1}{c}{29.63}  & 6.91                                      & \multicolumn{1}{c}{62.81}  \\
\textit{10x64x5x4 (S)}   &                  & \multicolumn{1}{l}{}                      &                            & \multicolumn{1}{l}{}                      &                            & \multicolumn{1}{l}{}                      &                            \\ \bottomrule
\end{tabular}
}
\end{table}

\begin{table}[h]
\centering
\caption{Evaluation of the capsule functions for RISC-V MCUs}
\label{tab:cap_risc_res}
\resizebox{0.8\linewidth}{!}{
\begin{tabular}{@{}llcccc@{}}
\toprule
\textbf{}                & \textbf{}                  & \multicolumn{2}{c}{\textbf{GAP-8 (single-core)}}                                      & \multicolumn{2}{c}{\textbf{GAP-8 (octa-core)}}                                        \\
\textbf{Dataset}         &                            & \multicolumn{2}{c}{\textit{RISC-V RV32IMCXpulp}}                                      & \multicolumn{2}{c}{\textit{RISC-V RV32IMCXpulp}}                                      \\ \cmidrule(l){3-6} 
                         &                            & \multicolumn{1}{l}{\textbf{Clock Cycles}} & \multicolumn{1}{l}{\textbf{Milliseconds}} & \multicolumn{1}{l}{\textbf{Clock Cycles}} & \multicolumn{1}{l}{\textbf{Milliseconds}} \\ \midrule
\textbf{MNIST}           & \textit{cap\_parallel\_q7} & 20.32M                                    & 119.52                                    & 7.96M                                     & 46.83                                     \\
\textit{10x1024x6x4 (L)} & \textit{}                  &                                           &                                           &                                           &                                           \\ \midrule
\textbf{smallNORB}       & \textit{cap\_parallel\_q7} & 16.26M                                    & 95.64                                     & 6.46M                                     & 38.03                                     \\
\textit{5x1600x6x4 (M)}  & \textit{}                  &                                           &                                           &                                           &                                           \\ \midrule
\textbf{CIFAR-10}        & \textit{cap\_parallel\_q7} & 4.55M                                     & 26.77                                     & 1.92M                                     & 11.28                                     \\
\textit{10x64x5x4 (S)}   & \textit{}                  &                                           &                                           &                                           &                                           \\ \bottomrule
\end{tabular}
}
\end{table}

\section{Related Work}
\citet{Zhang2020} proposed a hybrid computing architecture to accelerate the routing procedure of CapsNets during inference in GPU platforms. The authors claim that even in modern GPUs, CapsNets exhibit low efficiency due to their routing procedure. Through runtime profiling, \citet{Zhang2020} showed that the inefficiency of the routing procedure emerges from (i) tremendous data access to off-chip memory and (ii) intensive synchronizations to avoid race conditions on the limited GPU on-chip memory. To tackle this issue, the authors proposed a hybrid computing architecture named PIM-CapsNet. Pim-CapsNet offloads the routing procedure to an in-memory computation solution. \citet{Zhang2020} stated their hybrid computing architecture is 2.44x faster than a GPU-only execution. The previous work was later extended with software optimizations. After profiling, \citet{Zhang2021} stated that (i) nearby capsules have similar coupling coefficients (ii) and not all capsules require the same routing iterations to get stable coupling coefficients. \citet{Zhang2021} proposed then to group capsules in 3x3 groups, where only the central capsule is used in the computation of the coupling coefficients. To address the latter finding, \citet{Zhang2021} grouped capsules in two distinct groups: (i) one containing the capsules whose coupling coefficients are relatively stable and does not require the total routing iterations; (ii) and the other containing the capsules that benefit from the maximum number of iterations. With this software optimization, \citet{Zhang2021} achieved a significantly 3.41x speedup.

To counteract the intensive matrix multiplications and the complexity of the dynamic routing, \citet{Marchisio2021} proposed an RTL-level hardware architecture. At the core of their hardware accelerator, there is a processing element array, which is responsible for all matrix operations. This hardware accelerator also includes units specialized to speed up the \textit{ReLU}, \textit{squash}, and \textit{softmax} activations. Despite only supporting inputs as 8-bit integers, the bit-width of internal operations varies along the pipeline. To thread off the area, delay, and energy consumption of the accelerator, \citet{Marchisio2021} developed a framework that takes as input a given CapsNet model and a general architecture of the CapsNet accelerator and returns a set of optimal parameters for the accelerator. Results showed that the accelerator is 6x faster than a GPU-only solution (Nvidia GTX1070) \citep{Marchisio2021}. Based on this work, \citet{Park2020} proposed an accelerator enhanced for 3D CapsNets. This newer accelerator relies on a convolution core with 4 arrays of processing elements, each one designed for multiplications with 8-bit operands. For the design of the dynamic routing core, \citet{Park2020} stated that most of the coupling coefficients practically maintain their value between updates. The dynamic routing core was designed to skip the route operation when the variation is less than 0.005\%. These optimizations combined resulted in an accelerator that consumes 53.27\% less energy than the solution proposed by \citet{Marchisio2021}.

In an attempt to reduce the memory footprint of CapsNet, \citet{Marchisio2020} proposed a quantization framework. This framework explores different layer-wise and operation-wise precisions for obtaining a quantized CapsNet that trades off the classification accuracy and memory footprint. For this purpose, it takes as input a trained CapsNet model, a library of rounding schemes, an accuracy tolerance, and a maximum memory budget. For a CapsNet trained on the CIFAR-10 dataset, the framework reduces the memory footprint by 83\% with only 0.15\% accuracy loss.

\citet{Kakillioglu2020} proposed a weight pruning technique to reduce the large set of parameters in CapsNets based on dynamic routing, enabling their deployment in resource-constrained devices. Weight pruning techniques work by removing connections between neurons of a trained model. Weights in a tensor are ranked according to their magnitude and then the smallest of them, for a given percentage, are set to zero. The layer-wise pruning technique developed by \citet{Kakillioglu2020} enabled the deployment of CapsNets using 84.93\% to 97.01\% less memory than the original model proposed by \citet{Sabour2017}.

\subsection{Gap Analysis}
While previous studies accelerate the inference pass of CapsNets on GPUs, ASICs, and FPGA platforms, this paper goes a step further and evaluates the deployment of CapsNets on off-the-shelf Arm and RISC-V MCUs targeted for intelligent IoT nodes. We extended ready-available APIs (CMSIS-NN and PULP-NN) for quantized ANNs to support both primary capsule and capsule layers. As detailed in Table \ref{tab:Gap_Analysis}, while previous solutions propose hardware accelerators to tackle the heavy computation of CapsNets, we limit the computation spectrum to resource-constrained MCUs ready-available on the market. In addition to the price savings, this approach also has the potential to accelerate the adoption of CapsNets on the edge as it does not require novel and specific hardware solutions to be deployed. As our software kernels only support fixed 8-bit data, we developed a quantization framework. Due to constraints of CMSIS-NN and PULP-NN, our framework does not adopt a mixed bit-width quantization scheme as \citet{Marchisio2020}.

\begin{table}[b]
\caption{Functionality of previous research against our approach}
\label{tab:Gap_Analysis}
\resizebox{\linewidth}{!}{%
\begin{tabular}{@{}llll@{}}
\toprule
\textbf{R\&D Study}                      & \textbf{Target Platform}                           & \textbf{Problems Addressed}                              & \textbf{Proposal}                                                                                                                                                                                                        \\ \midrule
\citet{Zhang2020}       & GPU                                                & Latency                                                  & \begin{tabular}[c]{@{}l@{}}1. Hardware accelerator to offload the computation\\ of the dynamic routing from the GPU\end{tabular}                                                                                         \\ \midrule
\citet{Zhang2021}       & GPU                                                & Latency                                                  & \begin{tabular}[c]{@{}l@{}}1. Software optimization to optimize the dynamic routing\\ on an hardware accelerator coupled to a GPU\end{tabular}                                                                           \\ \midrule
\citet{Marchisio2021}   & FPGA                                               & \begin{tabular}[c]{@{}l@{}}Latency\\ Energy\end{tabular} & \begin{tabular}[c]{@{}l@{}}1. RTL-level hardware architecture for CapsNet inference\\ 2. Software to optimize the RTL architecture given a CapsNet\end{tabular}                                                          \\ \midrule
\citet{Park2020}        & ASIC                                               & \begin{tabular}[c]{@{}l@{}}Latency\\ Energy\end{tabular} & 1. RTL-level hardware architecture for 3D CapsNet inference                                                                                                                                                              \\ \midrule
\citet{Marchisio2020}   & \begin{tabular}[c]{@{}l@{}}GPU\\ FPGA\end{tabular} & Memory                                                   & 1. CapsNet quantization tool supporting mixed bit-width                                                                                                                                                                  \\ \midrule
\citet{Kakillioglu2020} & \begin{tabular}[c]{@{}l@{}}GPU\\ CPU\end{tabular}  & Memory                                                   & 1. Weight pruning tailored for CapsNets                                                                                                                                                                                  \\ \midrule
\textbf{Costa et. al}                    & MCU                                                & \begin{tabular}[c]{@{}l@{}}Memory\\ Latency\end{tabular} & \begin{tabular}[c]{@{}l@{}}1. CapsNet quantization tool supporting fixed int-8 data\\ 2. API for CapsNets inference in Armv7E-M, Armv8-M MCUs\\ 3. API for CapsNets inference in RISC-V (RV32IMCXpulp) MCUs\end{tabular} \\ \bottomrule
\end{tabular}
}
\end{table}

Future work may encompass the tune of our software kernels, along with the required kernels of
CMSIS-NN and PULP-NN, to support mixed bit-width. We believe that mixed bit-width quantization can further enhance our software kernels, as it gives more precision to weights with a higher impact on the final decision while reducing the precision of the others. This would be the perfect trade-off between memory footprint reduction and accuracy loss. Following the work from \citet{Kakillioglu2020}, we may also use a pruning scheme to enhance our quantization framework. This will also require the tune of our software kernels with techniques to optimize the loading and storing of zeroes in memory. Nevertheless, results detailed in Section \ref{sec:sw_kernels} show that our software kernels, at this stage of development, already achieve impressive gains in terms of memory footprint and latency, while relying on low-cost platforms available on the market. It is unfeasible to compare the latency speedup, memory footprint reduction, and accuracy loss of our solution with the others on Table \ref{tab:Gap_Analysis} as the baseline reference is not the same.

\section{Conclusion}
In this paper, we proposed the first public and open-source software kernels to enable the deployment of CapsNets in Arm Cortex-M and RISC-V MCUs. We extended the CMSIS-NN and PULP-NN to support the inference pass of CapsNets. As CapsNets are typically trained in 32-bit floating-point and low-end MCUs usually do not feature an FPU, we also proposed a framework, built upon TensorFlow, to quantize CapsNets to int-8 data.

Results show that our work succeeds in addressing the latency and memory footprint problems inherent to CapsNets, proving that it is possible to efficiently implement them on the deep edge. Our quantization framework reduces the memory footprint of a CapsNet by about 75\% with negligible accuracy loss. Furthermore, the latency induced by our software kernels on the execution of a primary capsule or capsule layer with medium-sized kernels can be lower than a few tenths of a second. To the best of our knowledge, this is the first work evaluating the deployment of quantized CapsNet on resource-constrained MCUs.

\bibliographystyle{unsrtnat}
\bibliography{ms.bib}  %%% Uncomment this line and comment out the ``thebibliography'' section below to use the external .bib file (using bibtex) .

%%% Uncomment this section and comment out the \bibliography{references} line above to use inline references.
% \begin{thebibliography}{1}

% 	\bibitem{kour2014real}
% 	George Kour and Raid Saabne.
% 	\newblock Real-time segmentation of on-line handwritten arabic script.
% 	\newblock In {\em Frontiers in Handwriting Recognition (ICFHR), 2014 14th
% 			International Conference on}, pages 417--422. IEEE, 2014.

% 	\bibitem{kour2014fast}
% 	George Kour and Raid Saabne.
% 	\newblock Fast classification of handwritten on-line arabic characters.
% 	\newblock In {\em Soft Computing and Pattern Recognition (SoCPaR), 2014 6th
% 			International Conference of}, pages 312--318. IEEE, 2014.

% 	\bibitem{hadash2018estimate}
% 	Guy Hadash, Einat Kermany, Boaz Carmeli, Ofer Lavi, George Kour, and Alon
% 	Jacovi.
% 	\newblock Estimate and replace: A novel approach to integrating deep neural
% 	networks with existing applications.
% 	\newblock {\em arXiv preprint arXiv:1804.09028}, 2018.

% \end{thebibliography}

\end{document}